\documentclass{article}
\usepackage[english]{babel}
\usepackage[utf8]{inputenc}
\usepackage{algorithm}
\usepackage{algpseudocode}
\usepackage[bbgreekl]{mathbbol}
\usepackage{amsmath}
\usepackage{amssymb}
\usepackage{bbm}
\usepackage{comment}
\usepackage{soul}
\usepackage[table, dvipsnames]{xcolor}
\usepackage{graphicx} 
\usepackage{booktabs} 
\usepackage{xparse}
\usepackage{array}
\usepackage{pifont}
\usepackage{booktabs}

\sethlcolor{yellow}
\NewDocumentCommand{\rot}{O{40} O{1em} m}{\makebox[#2][l]{\rotatebox{#1}{#3}}}
\newcolumntype{C}[1]{>{\centering\arraybackslash}p{#1}}
\DeclareMathOperator*{\argmax}{arg\,max}

\usepackage{johd}
\usepackage[citestyle=authoryear,bibstyle=apa,giveninits=true, maxnames=2, minnames=1, apamaxprtauth=99,backend=biber,uniquelist=false]{biblatex}
\AtEveryBibitem{\clearfield{month}}
\AtEveryBibitem{\clearfield{day}}
\AtEveryBibitem{\clearfield{language}}

\title{Going faster to see further: GPU-accelerated value iteration and simulation for perishable inventory control using JAX}

\author{Joseph Farrington$^{a}$$^{*}$, Kezhi Li$^{a}$, Wai Keong Wong$^{a,b}$, Martin Utley$^{c}$  \\
        \small $^{a}$Institute of Health Informatics, University College London, 222 Euston Road, London, NW1 2DA, UK \\
        \small $^{b}$NIHR University College London Hospitals Biomedical Research Centre, University College London and University\\  \small  College London Hospitals NHS Foundation Trust, 149 Tottenham Court Road, London, W1T 7DN, UK \\
        \small $^{c}$Clinical Operational Research Unit, University College London, 4 Taviton Street, London, NW1 2PG, UK \\\\
        \small $^{*}$Corresponding author: \tt{ucabjmf@ucl.ac.uk} \\
}
\date{}
\begin{document}
\maketitle
\begin{abstract} 
\noindent Value iteration can find the optimal replenishment policy for a perishable inventory problem, but is computationally demanding due to the large state spaces that are required to represent the age profile of stock.  The parallel processing capabilities of modern GPUs can reduce the wall time required to run value iteration by updating many states simultaneously.  The adoption of GPU-accelerated approaches has been limited in operational research relative to other fields like machine learning, in which new software frameworks have made  GPU programming widely accessible. We used the Python library JAX to implement value iteration and simulators of the underlying Markov decision processes in a high-level API, and relied on this library's function transformations and compiler to efficiently utilize GPU hardware. Our method can extend use of value iteration to settings that were previously considered infeasible or impractical. We demonstrate this on example scenarios from three recent studies which include problems with over 16 million states and additional problem features, such as substitution between products, that increase computational complexity. We compare the performance of the optimal replenishment policies to heuristic policies, fitted using simulation optimization in JAX which allowed the parallel evaluation of multiple candidate policy parameters on thousands of simulated years. The heuristic policies gave a maximum optimality gap of 2.49\%. Our general approach may be applicable to a wide range of problems in operational research that would benefit from large-scale parallel computation on consumer-grade GPU hardware. \end{abstract}

\noindent\\

\section{Introduction}

Perishable items, such as fresh food and blood products,  ``\textit{undergo change in storage so that in time they may become partially or entirely unfit for consumption}'' \parencite{Nahmias1982}. This means that wastage must be considered alongside the impact of shortages and stock-holding levels when making replenishment decisions. Wastage may be a concern for economic, sustainability, or ethical reasons. One target of the United Nations Sustainable Development Goals is to reduce wastage at the retail and consumer levels of the food supply chain by half, from an estimated 17\% of total food production \parencite{united_nations_sustainable_2022}. In the blood supply chain, platelets can only be stored for between three and seven days leading to high reported wastage rates of 10-20\% \parencite{flint_is_2020}. Better policies for perishable inventory control could help to reduce wastage and make the best possible use of limited resources. 

Early theoretical work demonstrated that optimal policies for perishable inventory replenishment could be found using dynamic programming \parencite{nahmias_optimal_1975, fries_optimal_1975}. Value iteration is a dynamic programming approach that can be used to find the optimal policy when the problem is framed as a Markov decision process \parencite{bellman_dynamic_1957}. The optimal policy depends on the age profile of the inventory, not just on the total number of units in stock. This approach is therefore limited by the ``curse of dimensionality'': the computational requirements grow exponentially with the maximum useful life of the product \parencite{nahmias_perishable_2011}. \textcite{Nahmias1982} observed that, at the time, this made dynamic programming approaches impractical for problems where the maximum useful life of the product was more than two periods.  More recently, despite advances in computational power, researchers have stated that value iteration remains infeasible or impractical when the maximum useful life of the product is longer than two or three periods or when additional complexities (e.g. substitution between products or random remaining useful life on arrival) are introduced \parencite{de_moor_reward_2022, hendrix_computing_2019, mirjalili_data-driven_2022}

The prevalent view in the literature about the scale of problems for which value iteration is feasible appears to neglect the recent developments in graphical processing unit (GPU) hardware, and in software libraries that make it possible for researchers and practitioners to take advantage of GPU capabilities without detailed knowledge of GPU hardware and GPU-specific programming approaches. GPUs were developed for rendering computer graphics, which requires the same operations to be efficiently applied to many inputs in parallel. Compared to a central processing unit (CPU), GPUs therefore have many more, albeit individually less powerful, cores. GPU-acceleration refers to offloading computationally intensive tasks that benefit from large-scale parallelization from the CPU to the GPU. The first mainstream software framework to support general computing tasks on GPUs using a common general purpose programming language was Nvidia's CUDA platform, which launched in 2007. One of the areas in which GPUs have since had a major impact is the field of deep learning. This impact has led to, and in turn been supported by \parencite{jeon_chapter_2021}, the development of higher-level software libraries  including TensorFlow \parencite{Abadi2016}, PyTorch \parencite{paszke_pytorch_2019} and JAX \parencite{bradbury_jax_2018}. These libraries provide comparatively simple Python application programming interfaces (APIs) to support easy experimentation and developer productivity, while utilising highly optimized CUDA code ``under-the-hood'' to exploit the parallel processing capabilities of GPUs.

In this work, we implemented value iteration using JAX. JAX provides composable transformations of Python functions, which make it easy to apply functions in parallel over arrays of input data, and performs just-in-time compilation (JIT) to run workloads efficiently on hardware accelerators including GPUs. This is ideal for value iteration: each iteration requires many independent updates which can be performed in parallel, and the up-front computational cost of JIT can be amortised over many repeats of the compiled operation for each iteration. By decreasing the wall time required to run value iteration, we increase the size of problems for which the optimal policy can be calculated in practice: by going faster, we can see further. These policies may themselves be used to guide decision making. They can also support research into new heuristics and approximate approaches, including reinforcement learning, by providing performance benchmarks on much larger problems than has previously been possible.

Inspired by the recent development of GPU-based simulators in reinforcement learning \parencite{freeman_brax_2021, makoviychuk_isaac_2021, lange_gymnax_2022, bonnet_jumanji_2022}, we also implemented simulators for perishable inventory problems using JAX, which enabled us to run large numbers of simulations in parallel.

We consider perishable inventory scenarios from three recent studies, where running value iteration for certain settings was described as computationally infeasible or impractical. For most of these settings, we have been able to find the optimal policy using a consumer-grade GPU and report the wall time required for each experiment. We compare the performance of the policies found using value iteration with the performance of heuristic policies with parameters fitted using simulation optimization. 

The main contributions of this work are in demonstrating that:
\begin{itemize}
\item{ value iteration can be used to find optimal policies, for scenarios for which it has recently been described as computationally infeasible or impractical, on consumer-grade GPU hardware (summarised in Table \ref{tab:intro:contrib:m_table});}
\item{this performance can be achieved without in-depth knowledge of GPU-specific programming approaches and frameworks, using the Python library JAX;}
\item{simulation optimization for perishable inventory control can also be effectively run in parallel using JAX, particularly for larger problems where policies that perform well can be identified in a fraction of the time required to run value iteration.}
\end{itemize}

For one large problem with over 16 million states, a CPU-based MATLAB implementation of value iteration did not converge within a week in a prior study. Using our method, value iteration converges in under 3.5 hours on a consumer-grade GPU and, without any code changes, in less than 30 minutes using four data-centre grade GPUs. Our simulation optimization method is able to evaluate 50 possible sets of parameters for a heuristic policy, each on 4,000 years of simulated data, in parallel in under 15 seconds. The largest optimality gap we observed for the heuristic policies fit using our simulation optimization method was 2.49\%. 

Our open-source code is available at \url{https://github.com/joefarrington/viso_jax}. The repository includes a Google Colab notebook that enables interested readers to reproduce our experiments using free cloud-based GPUs provided by Google. 

\begin{table}[h!]
\centering
\begin{tabular}{ll*{5}{C{0.4cm}}}
\toprule
\multicolumn{2}{l}{} & \multicolumn{5}{c}{\textbf{Maximum useful life} $m$} \\
\cmidrule(lr){3-7}
& Problem features & 2 & 3 & 4 & 5 & 8 \\
\midrule
A & Lead time $>$ 1 & \textcolor{blue} {\ding{108}} & \textcolor{ForestGreen}{\ding{81}} & \textcolor{ForestGreen}{\ding{81}} & \textcolor{ForestGreen}{\ding{81}} & \textcolor{gray}{\ding{109}} \\
&  & &  &  &  \\
B & Substitution between products & \textcolor{blue} {\ding{108}} & \textcolor{ForestGreen}{\ding{81}} & \textcolor{gray}{\ding{109}} & \textcolor{gray}{\ding{109}} & \textcolor{gray}{\ding{109}} \\
&  &  &  &  &  \\
C & Not all arrivals fresh, periodic demand & \textcolor{gray}{\ding{109}} & \textcolor{blue} {\ding{108}} & \textcolor{gray}{\ding{109}} & \textcolor{ForestGreen}{\ding{81}} & \textcolor{OrangeRed}{\ding{54}} \\
\bottomrule
\end{tabular}
\quad
\begin{tabular}{c l}
& \\
& \textbf{Key} \\
   \textcolor{blue} {\ding{108}} & Value iteration feasible for all experiments in the original study. \\
\textcolor{ForestGreen}{\ding{81}} & Our method extends value iteration to experiments that were considered  \\
& infeasible or impractical in the original study. \\
\textcolor{OrangeRed}{\ding{54}} & All experiments infeasible in the original study and with our method. \\
 \textcolor{gray}{\ding{109}} & Setting not considered in the original study. \\
\end{tabular}
\caption{\label{tab:intro:contrib:m_table}Summary of our contribution extending value iteration to larger problems with a longer \\maximum useful life.}
\end{table}

\section{Related work}

\textcite{johannsson_gpu-based_2009} demonstrated that value iteration could be effectively run in parallel on a GPU soon after the introduction of CUDA. Subsequent research has evaluated the performance of GPU-accelerated value iteration on problems from economics and finance \parencite{aldrich_tapping_2011, aamer_data_2020, duarte_benchmarking_2020, kirkby_toolkit_2017, kirkby_quantitative_2022} and route-finding and navigation \parencite{chen_markov_2013, inamoto_implementation_2011, ruiz_parallel_2015, constantinescu_performance_2020}. We have only identified a single study that applied this approach to an inventory control problem: \textcite{ortega_cuda_2019} implemented a custom value iteration algorithm in CUDA to find replenishment policies for a subset of perishable inventory problems originally described by \textcite{hendrix_computing_2019}.

Previous studies have reported impressive reductions in wall time achieved by running value iteration on GPU. The GPU-accelerated method in \textcite{ortega_cuda_2019} was up to 11.7$\times$ faster than a sequential CPU-based method written in C. Despite this, the approach has not been widely adopted. The majority of the aforementioned studies focus on implementing value iteration using CUDA or OpenCL, a multi-platform alternative to CUDA, and comparing the performance of a GPU-accelerated method with a CPU-based method, instead seeking to use GPU-acceleration to solve problems for which value iteration is otherwise impractical or infeasible. One of the main barriers to entry for other researchers may be the perceived difficulty of GPU programming. Writing efficient code using CUDA or OpenCL requires careful consideration of memory access, balancing resource usage when mapping parallel processes to the hardware, and interaction between the CPU and GPU \parencite{hijma_optimization_2022}. There have been efforts to make GPU-accelerated value iteration more accessible. \textcite{johannsson_gpu-based_2009} created a solver framework using his CUDA implementation of value iteration as a back-end, but this does not appear to be publicly available. More recently, \textcite{kirkby_toolkit_2017} created a toolkit in MATLAB to solve infinite horizon value iteration problems which automatically uses a GPU when available and appropriate. This toolkit requires the state-transition matrix to be provided as an input, which is not practical for some of the larger problems we consider due to memory limitations. 

Our approach is broadly similar to that of \textcite{duarte_benchmarking_2020} and \textcite{sargent_dynamic_2022} who used machine learning frameworks to implement GPU-accelerated value iteration for economics models. A key observation made by \textcite{duarte_benchmarking_2020} is that their TensorFlow implementation is an order of magnitude faster than their custom CUDA C++ implementation on a GPU. This demonstrates that the comparative accessibility provided by a machine learning framework need not come at the cost of poorer performance. TensorFlow, like PyTorch and JAX, translates the high level API instructions into highly-optimized CUDA code. Experts in GPU programming may be able achieve better performance than a machine learning framework by working at the level of CUDA or OpenCL. Researchers without that expertise are likely to both save development time and achieve performance benefits by working at a higher level of abstraction using a machine learning framework and relying on it to make best use of the available hardware. 

Due to the computational challenges of using dynamic programming methods to find policies for perishable inventory management, research has focused on approximate solutions. One straightforward way to make a dynamic programming approach more computationally tractable is to reduce the size of the state space by aggregating stock items into batches. The solution to the down-sized dynamic program can then by factored up to give an approximate solution to the the original problem (e.g. \cite{blake_optimizing_2003, haijema_blood_2007}). An alternative approach is to use a heuristic policy with a small number of parameters, such as a base-stock policy. Research in this area has concentrated on both identifying suitable structures for heuristic policies (see \textcite{nahmias_comparison_1975} for an early example, and \textcite{haijema_improved_2019} for a recent example), and finding suitable parameters for those policies in specific situations - commonly using stochastic mixed integer linear programming (e.g. \cite{Dillon2017, Gunpinar2015, rajendran_platelet_2017}) or simulation optimization (e.g. \cite{dalalah_platelets_2019, duan_new_2013}). Recently, reinforcement learning methods have also been used to find approximate polices for managing perishable inventory \parencite{Kara2018a, Sun2019, de_moor_reward_2022, ahmadi_intelligent_2022}. 

Of these other approaches to the problem, we focus on simulation optimization in addition to value iteration because GPU-accelerated simulation is feasible using available software libraries but not yet widely adopted. Applied research on specific mixed integer linear programs often relies on commercial solver software such as IBM ILOG CPLEX Optimization Studio or Gurobi Optimizer which do not currently support GPU-acceleration. Adapting mixed integer programming solution strategies to suit the architecture of GPUs is the subject of active research \parencite{perumalla_design_2021}. The use of GPU-acceleration supported by machine learning frameworks is already widespread in reinforcement learning research. 

Simulation optimization can be used to solve optimization problems where the objective function cannot be computed exactly, but can be estimated using simulation. Sampling error in the objective function can be reduced by running simulations for a longer period, or by running additional simulations. The relevance of parallel computing to simulation optimization is well recognised \parencite{amaran_simulation_2016, fu_simulation_2014}, but we have identified few simulation optimization studies using GPUs to run multiple simulations in parallel. In inventory management, \textcite{srimool_speeding_2011} exhaustively evaluated the possible order quantities for a newsvendor problem using parallel simulations on GPU. More recently, \textcite{lau_hybrid_2016} used simulation optimization to solve a chemical process monitoring problem using GPU-acceleration for both the simulation and the metaheuristic search process that proposed candidate solutions. Similar to the value iteration studies discussed above, both of these projects used custom CUDA code which may explain the limited subsequent adoption despite the established reductions in wall time relative to CPU baselines. Following recent work in the reinforcement learning community \parencite{freeman_brax_2021,lange_evosax_2022} we implemented our simulators using the Python library gymnax \parencite{lange_gymnax_2022}, which enabled us to write our simulation operations using JAX and readily evaluate each of numerous policies on thousands of simulated years in parallel on GPU. 

\section{Methods}

\subsection{Scenarios}

We considered three scenarios, all of which are periodic review, single-echelon perishable inventory problems with a fixed, known delivery lead time $L$. The three scenarios were selected as recent examples from the perishable inventory literature in which value iteration was reported as infeasible or impractical for at least some experimental settings, and which include elements relevant to our wider work investigating the potential of reinforcement learning methods to support blood product inventory management. Scenario A, from \textcite{de_moor_reward_2022}, is a straightforward perishable inventory replenishment problem but for some experimental settings the lead time, $L$, is greater than one period and therefore we need to consider inventory in transit when placing an order. Scenario B is the two product scenario described by \textcite{hendrix_computing_2019} which adds the complexity of substitution between perishable products. Substitution is an important aspect of managing blood product inventory, where compatibility between the blood groups of the donor and the recipient is critical. Scenario C, from \textcite{mirjalili_data-driven_2022}, models the management of platelets in a hospital blood bank and adds two complicating factors: periodic patterns of demand, and uncertainty in the remaining useful life of products on arrival, which may depend on the order quantity. In every scenario demand is stochastic, unmet demand is assumed to be lost, and units in stock with a remaining useful life of one period are assumed to expire at the end of the day. Except in Scenario C, the products have a fixed, known useful life $m$ and are all assumed to arrive fresh. We summarise the key differences between the scenarios in Table \ref{tab:scenarios}.

For readers who are unfamiliar with inventory management problems and the associated terminology we recommend Chapters 3 and 4 of \textcite{snyder_fundamentals_2019} for a general introduction and \textcite{chaudhary_state---art_2018} and \textcite{nahmias_perishable_2011} for more focused coverage of perishable inventory control. 

\begin{table}[h!]
\centering
\footnotesize
\begin{tabular}{ll *{14}{C{0.4cm}}C{0.2cm}}
\toprule
\multicolumn{2}{c}{} & \multicolumn{6}{c}{\bf{Problem features}} & \multicolumn{2}{c}{} & \multicolumn{6}{c}{\bf{Reward function components}} &\\
\cmidrule(lr){3-8} \cmidrule{11-16}
& Source & \rot{Products} & \rot{Lead time $> 1$} & \rot{Substitution} & \rot{Not all arrivals fresh} & \rot{Periodic demand} & & & &\rot{Variable ordering} & \rot{Fixed ordering} & \rot{Wastage} & \rot{Shortage} & \rot{Holding} & \rot{Revenue} \\
\midrule
A & \textcite{de_moor_reward_2022} & 1 & \checkmark & & & & &&& \checkmark & & \checkmark & \checkmark & \checkmark &\\
&  & &  &  &  & &  &  &  &  \\
B & \textcite{hendrix_computing_2019} & 2 &  & \checkmark & & &&&  & \checkmark & & & & & \checkmark\\
&  & &  &  &  & & &&& &  &  &  \\
C & \textcite{mirjalili_data-driven_2022} & 1 & & & \checkmark & \checkmark &  & &&& \checkmark & \checkmark & \checkmark & \checkmark &\\
\bottomrule
\end{tabular}
\caption{\label{tab:scenarios}Summary of the key differences between our three scenarios.}
\end{table}

\textcite{de_moor_reward_2022}, \textcite{hendrix_computing_2019} and \textcite{mirjalili_data-driven_2022} each used different notation to describe their work. In an effort to aid the reader in understanding the similarities and differences between the scenarios we have adopted a single notation which we apply to all three scenarios. In Appendices \ref{appendix:a:desc}, \ref{appendix:b:desc} and \ref{appendix:c:desc} we present the key equations describing each scenario in our notation and provide a table summarising our notation in Appendix \ref{appendix:notation}.

All of the scenarios are defined as Markov decision processes (MDPs). An MDP is a formal description of a sequential decision problem in which, at a discrete series of points in time, an agent observes the state of its environment $S_t$ and selects an action $A_t$. At the next point in time, the agent will receive a reward signal $R_{t+1}$, observe the updated state of its environment $S_{t+1}$ and must select its next action $A_{t+1}$. An MDP can be defined in terms of a set of states $s \in \mathbb{S}$, a set of actions $a \in \mathbb{A}$, a set of a rewards $r \in \mathbb{\Psi}$, a function defining the dynamics of the MDP (Equation \ref{eq:mdp_dynamics}), and a discount factor $\gamma \in [0,1]$ \parencite{sutton_reinforcement_2018}. The discount factor controls the relative contribution of future rewards and immediate rewards. The decision process is Markovian because the dynamics of the system obey the Markov property: state transitions and rewards at time $t$ are conditionally independent of the sequence of state-action pairs $(S_0, A_0)$ to $(S_{t-1}, A_{t-1})$ given $(S_t, A_t)$. 

\begin{equation} \label{eq:mdp_dynamics}
p(s', r|s,a) = \text{Prob}\left(S_t=s', R_t=r|S_{t-1}=s, A_{t-1}=a\right)
\end{equation}

Within this framework, MDP agents select their actions by following a policy. In this work, we only consider deterministic policies, $a=\pi(s)$. The objective is to find a policy that maximises the expected return, the discounted sum of future rewards, when interacting with the environment. In an infinite horizon problem the return at timestep $t$ is $G_t = \sum_{k=0}^{\infty} \gamma^{k}R_{t+k+1}$. 

\subsection{Value iteration}

Following the treatment of \textcite{sutton_reinforcement_2018}, the value of a state under a policy $\pi$, $V^{\pi}(s)$, is the expected return when starting in state $s$ and following policy $\pi$. For a finite MDP, one in which the sets of states, actions and rewards are finite, we can define an optimal policy, $\pi^*$, as a policy for which $V^{\pi^*}(s) \geq V^{\pi'}(s)$ for every state $s \in \mathbb{S}$, for any policy $\pi'$. There may be more than one optimal policy, but they all share the same optimal value function. Value functions satisfy recursive relationships, called Bellman equations, between the value at the current state and the immediate reward plus the discounted value at the next state. The Bellman equation for the optimal policy, the Bellman optimality equation, is:

\begin{equation} \label{eq:bellmanoptv}
    V^{\pi^*}(s) = \max_{a \in \mathbb{A}} \sum_{s' \in \mathbb{S}, r \in \mathbb{\Psi}} p(s',r|s,a) \left[r + \gamma V^{\pi^*}(s')\right]
\end{equation}

Value iteration is a dynamic programming algorithm, which uses the Bellman optimality equation as an update operation to estimate the optimal value function:

\begin{equation} \label{eq:valueiterationupdate}
    V_{i+1}(s) = \max_{a \in \mathbb{A}} \sum_{s' \in \mathbb{S}, r \in \mathbb{\Psi}} p(s',r|s,a) \left[r + \gamma V_i(s')\right]
\end{equation}

For a finite MDP this operation will, in the limit of infinite iterations, converge to the optimal value function. The optimal policy can be extracted from the value function using a one-step ahead search:
\begin{equation} \label{eq:valueiterationpolicy}
\pi(s) = \argmax_{a \in \mathbb{A}} \sum_{s' \in \mathbb{S}, r \in \mathbb{\Psi}} p(s',r|s,a) \left[r + \gamma V(s')\right]
\end{equation}

Similar to the approach of \textcite{hendrix_computing_2019}, we used a deterministic transition function, $(s', r) = T(s, a, \omega)$, where $\omega \in \mathbb{\Omega}$ is a possible realisation of the stochastic element(s) of the transition. For a specific state-action pair, $(s,a)$, and a specific random outcome, $\omega$, the next state and the reward can be calculated deterministically. In the most straightforward example we consider, Scenario A, the only uncertainty is in the daily demand, and therefore $\mathbb{\Omega}$ is the set of possible values that demand may take in any period. Under this formulation, the value iteration update equation can be rewritten as:

\begin{equation} \label{eq:valueiterationdetupdate}
    V_{i+1}(s) = \max_{a \in \mathbb{A}} \sum_{\omega \in \mathbb{\Omega}} P(\omega|s,a) \left[r_{\omega} + \gamma V_i(s'_{\omega})\right], \text{where } (r_{\omega}, s'_{\omega}) = T(s, a, \omega)
\end{equation}

\noindent and we extract the optimal policy using the equation:

\begin{equation} \label{eq:valueiterationdetpolicy}
    \pi(s) = \argmax_{a \in \mathbb{A}} \sum_{\omega \in \mathbb{\Omega}} P(\omega|s,a) \left[r_{\omega} + \gamma V(s'_{\omega})\right], \text{where } (r_{\omega}, s'_{\omega}) = T(s, a, \omega)
\end{equation}

\noindent where $P(\omega|s,a) = \text{Prob}\left(\Omega_t = \omega | S_t = s, A_t = a\right)$ is the probability of random outcome $\omega$ having observed state $S_t=s$ and then taken action $A_t=a$. $\Omega_t$ represents the stochastic elements of the transition that occur between the observation of state $S_t$ and the observation of state $S_{t+1}$. 

Since we cannot run an infinite number of iterations, we use a convergence test to determine when to stop value iteration and extract the policy. We are interested in the policy, and not the value function itself, and therefore in certain cases we can reduce the number of iterations required by stopping when further updates to the value function will not change the policy. We describe the convergence test used for each scenario in the corresponding section below. 

We implemented value iteration using a custom Python class, \texttt{VIRunner}, which defines the common functionality required to run value iteration and extract the optimal policy using the approach in Equations \ref{eq:valueiterationdetupdate} and \ref{eq:valueiterationdetpolicy}. The base \texttt{VIRunner} class includes eight placeholders for methods which must be defined for a specific scenario. For each scenario we defined a subclass of \texttt{VIRunner}, replacing the placeholder methods with custom functions that:

\begin{itemize}
\item{return a list of all possible states as tuples;}
\item{return an array that maps from a state to its index in the list of all possible states;}
\item{return an array of all possible actions;}
\item{return an array of all possible random outcomes;}
\item{return the immediate reward and next state following the deterministic transition function given a state, action and random outcome;}
\item{return an array with the probability of each random outcome given a state-action pair;} 
\item{return an initial estimate of the value function; and}
\item{test for convergence of the value iteration procedure.}
\end{itemize}

The \texttt{VIRunner} class could be easily adapted to solve new problems by creating a new subclass and replacing the placeholder entries for these eight methods.  

A naive implementation of value iteration following Equation \ref{eq:valueiterationdetupdate} would require a nested for-loop over every state, every action, and every random outcome. In a single iteration, these updates are independent and therefore can be performed in parallel. JAX provides two main composable function transformations that facilitate running functions in parallel: vectorizing map (vmap) and parallel map (pmap). Both of these transformations create new functions that map the original function over specified axes of the input, enabling the original function to applied to a large number of inputs in parallel. The key difference is that vmap provides vectorization, so the operations happen on the same device (e.g. the same GPU), while pmap supports single-program, multiple-data parallelism and runs the operation for different inputs on separate (but identical) devices.

An important feature of vmap and pmap is that they are composable and therefore can be readily nested. For our basic value iteration update, set out in Algorithm \ref{alg:viter}, we nest vmap operations over states, actions, and random outcomes instead of using nested loops. This is only feasible if there is sufficient GPU memory to update all of the states simultaneously. For larger instances we grouped the states into batches for which the update can be performed simultaneously and performed the update for one batch of states at a time. To enable multiple identical devices to be used where available, we automatically detected the the number of available of devices and used pmap to map our update function for multiple batches of states over the available devices. Each batch must contain the same number of states, and each device must receive the same number of batches, to efficiently loop over batches of states and use pmap. We therefore padded the array of states so that it could be reshaped to an array with dimensions (\textit{number of devices}, \textit{number of batches}, \textit{maximum batch size}, \textit{number of elements in state}). Each device received an array with dimensions (\textit{number of batches}, \textit{maximum batch size}, \textit{number of elements in state}), performed a loop over the leading dimension, and calculated the update one batch of states at a time. The same process was used to extract the policy in parallel at the end of value iteration. 

\begin{algorithm}[ht]
\caption{Value iteration using vmap}\label{alg:viter}
\begin{algorithmic}
\State Initialise array of all states $s$: $\mathbb{S}$ 
\State Initialise array of all actions $a$: $\mathbb{A}$ 
\State Initialise array of all random outcomes $\omega$: $\mathbb{\Omega}$
\State Initialise initial estimate of value function: $V_0(s) \quad \forall s \in \mathbb{S}$
\State Initialise discount factor: $\gamma$
\State Initialise iteration counter: $i = 0$
\State Define deterministic transition function which returns next state and reward: $T(s, a, \omega)$
\\
\State \textbf{Perform value iteration}
\While{not converged}
\State $i \gets i+1$
\State \textbf{vmap} over $s \in \mathbb{S}$
\State \quad \textbf{vmap} over $a \in \mathbb{A}$
\State \quad \quad \textbf{vmap} over $\omega \in \mathbb{\Omega}$
\State \quad \quad \quad $(s'_{\omega}, r_{\omega}) \gets T(s, a, \omega)$
\State \quad \quad $Q_i(s, a) \gets \sum_{\omega} P(\omega|s,a) \left[r_{\omega} + \gamma V_{i-1}(s'_{\omega})\right]$
\State \quad $V_i(s) \gets \max_a Q_i(s,a)$
\State Test for convergence
\EndWhile
\\ 
\State \textbf{Extract the policy, $\pi(s) \approx \pi^*(s)$}
\State \textbf{vmap} over $s \in \mathbb{S}$
\State \quad \textbf{vmap} over $a \in \mathbb{A}$
\State \quad \quad \textbf{vmap} over $\omega \in \mathbb{\Omega}$
\State \quad \quad \quad $(s'_{\omega}, r_{\omega}) \gets T(s, a, \omega)$
\State \quad \quad $Q_{i+1}(s, a) \gets \sum_{\omega} P(\omega|s,a) \left[r_{\omega} + \gamma V_{i}(s'_{\omega})\right]$
\State \quad $\pi(s) \gets \argmax_a Q_{i+1}(s,a)$
\State
\end{algorithmic}
\end{algorithm}

Functions transformed by pmap are automatically JIT compiled with XLA, a domain-specific compiler for linear algebra. JAX traces the function the first time it is run, and the traced function is compiled using XLA into optimized code for the available devices.

 We used double-precision (64-bit) numbers when running value iteration, instead of the single-precision (32-bit) numbers that JAX uses by default because, in preliminary experiments, we found that convergence was not always stable. 

We report the wall time required to run each value iteration experiment. The reported times include JIT compilation time, writing checkpoints and writing final outputs, including the policy, because we believe this represents a realistic use case. 

\subsection{Simulation of the Markov decision processes}

We created a simulator to represent each scenario. The simulators have two purposes: firstly, to fit parameters for heuristic replenishment policies using simulation optimization and, secondly, to evaluate the performance of the policies produced by value iteration and simulation optimization based on the return and three key performance indicators (KPIs): service level, wastage and holding. The service level is the percentage of demand that was met over a simulated rollout, wastage is the proportion of units received that expired over a simulated rollout and holding is the mean number of units in stock at the end of each day during a simulated rollout. 

Each simulator is a reinforcement learning environment written using the Python library gymnax \parencite{lange_gymnax_2022}, which is based on JAX. This provides a standard interface for working with MDPs, while enabling many simulations to be run in parallel on a GPU using vmap. For a single policy, we can use vmap to implement our simulation rollout over multiple random seeds. We can simultaneously evaluate multiple sets of parameters for the same policy on a shared set of random seeds by nesting vmapped functions. We note that it would be straightforward to use reinforcement learning software libraries to learn policies for these scenarios using these environments. 

We selected different heuristic policies for the different scenarios from the literature, considering which (if any) heuristic was used in the original study and the structure of each problem. We describe the heuristic policy used for each scenario in the corresponding section below. All of the heuristic policies use one or both of an order-up-to level parameter \texttt{S} and reorder point parameter \texttt{s}. The order quantity is the difference between the current stock on hand and in transit (potentially subject to some modification, as in Scenario B) and the order-up-to level \texttt{S}. If there were no stock on hand or in transit the heuristic policy would order \texttt{S} units, and therefore \texttt{S} corresponds to the largest order that would be placed following the heuristic policy. If the heuristic policy also has a reorder point parameter \texttt{s} then an order is only placed when the current stock on hand and in transit is less than or equal to the reorder point \texttt{s} \parencite{snyder_fundamentals_2019}. 

We used the Python library Optuna \parencite{akiba_optuna_2019} to suggest parameters for the heuristic policies. For heuristic policies with a single parameter, we evaluated all feasible values simultaneously in parallel using Optuna's grid sampler. When there was more than one parameter we instead used a genetic algorithm, Optuna's NSGAII sampler, to search the parameter space. For each suggested set of parameters, we ran 4,000 rollouts, each 365 days long following a warm-up period of 100 days. When using the grid search sampler, we took as the best parameter value the one with the highest mean return after the single parallel run. When using the NSGAII sampler we ran 50 sets of parameters in parallel, representing a single generation for the genetic algorithm, and ranked them based on the mean return. We terminated the NSGAII search procedure when the best combination of parameters had not changed for five generations, or when 100 generations had been completed.  

For each scenario we compare the performance of the value iteration policy and best heuristic policy identified using simulation optimization on 10,000 simulated rollouts, each 365 days long following a warm-up period of 100 days. We report the mean and standard deviation of the return and, in the appendices, the service level, wastage and stock holding over these rollouts. For each rollout, the return is the discounted sum of rewards from the end of the warm-up period until the end of the simulation. The components of the reward function for Scenarios A, B and C are summarised in Table \ref{tab:scenarios} and the reward functions are set out in Appendices \ref{appendix:a:desc}, \ref{appendix:b:desc} and \ref{appendix:c:desc} respectively. The standard deviation of the return and the KPIs shows the effect of the stochasticity (due to random demand, random willingness to accept substitution and/or random useful life on arrival) in each scenario. 

\subsection{Reproducibility}

There are two key reproducibility considerations for this work: firstly, accurately implementing the scenarios described in previous studies and, secondly, ensuring that others are able to reproduce our own experiments.  

We compared outputs from our value iteration and simulation optimization methods to outputs from the original studies, and these checks are included as automated tests in our publicly available GitHub repository. \textcite{de_moor_reward_2022} made their code available on GitHub and fully specified the optimal and heuristic policies for two experiments in their paper which we used to test our implementation of Scenario A. For Scenario B, we compared the best parameters for heuristic policies and mean daily reward values to those reported in \textcite{hendrix_computing_2019}, and performed additional comparisons to the output of a MATLAB implementation of their value iteration method that the authors kindly made available to us. \textcite{mirjalili_data-driven_2022} plotted value iteration policies for a subset of his experiments, and he kindly provided us with the underlying data for those plots so that we could confirm the policies from our implementation of Scenario C matched those he had reported. 

Our code is available on GitHub, and is based on open-source software libraries. Our GitHub repository includes a Google Colab notebook that can be used to reproduce our experiments using a free, cloud-based GPU, avoiding local hardware requirements or configuration challenges. The type of GPU allocated to a session in Colab is not guaranteed and there are service limits that restrict the maximum continuous running time. Experiments may be restarted from a checkpoint if a session terminates before the experiment is completed. 

\subsection{Hardware}

All experiments were conducted on a desktop computer running Ubuntu 20.04 LTS via Windows Subsystem for Linux on Windows 11 with an AMD Ryzen 9 5900X processor, 64GB RAM, and an Nvidia GeForce RTX 3060 GPU. The Nvidia GeForce RTX 3060 is a consumer-grade GPU that, at the time of writing in February 2023, can be purchased for less than £400 in the United Kingdom \parencite{ebuyer_nvidia_nodate}.

To demonstrate the potential benefits of more powerful data-centre grade GPU devices, and how our approach can be easily scaled to utilise multiple GPUs, we additionally ran value iteration for one large problem case of Scenario B using one, two or four Nvidia A100 40GB GPUs.

\section{Scenario A: lead time may be greater than one period} \label{s:de_moor}

\subsection{Problem description}

\textcite{de_moor_reward_2022} described a single-product, single-echelon, periodic review perishable inventory replenishment problem and investigated whether using heuristic replenishment policies to shape the reward function can improve the performance of reinforcement learning methods.

At the start of each day $t$ the agent observes the state $S_t$, the current inventory in stock (split by remaining useful life) and in transit (split by period ordered), and places a replenishment order $A_t \in \{0, 1, ..., A_{\max}\}$. Demand for day $t$, $D_t$, is sampled from a truncated gamma distribution and rounded to the nearest integer. Demand is filled from available stock following either a first-in first-out (FIFO) or last-in first-out (LIFO) issuing policy. At the end of the day, the state is updated to reflect the ageing of stock and the reward, $R_{t+1}$, is calculated. The reward function comprises four components: a holding cost per unit in stock at the end of the period ($C_h$), a variable ordering cost per unit ($C_v$), a shortage cost per unit of unmet demand ($C_s$) and a wastage cost per unit that perishes at the end of the period ($C_w$). The order placed on day $t-(L-1)$ is received immediately prior to the start of day $t+1$, and is included in the stock element of the state $S_{t+1}$. 

The stochastic element in the transition is the daily demand $D$, $\mathbb{\Omega} =  \{0, 1, ..., \infty\}$, in the problem described by \textcite{de_moor_reward_2022}. The state transition and the reward are deterministic given a state-action pair and the realisation of the daily demand. Daily demand is modelled by a gamma distribution with mean $\mu$ and coefficient of variation $\frac{\mu}{\sigma}$. We truncated the demand distribution at $D_{\max} \gg \mu+5\sigma$, such that $\mathbb{\Omega} =  \{0, 1, ..., D_{\max}\}$, for the purposes of implementation. 

The initial value function $V_0(s)$ was initialised at zero for every state. \textcite{de_moor_reward_2022} did not specify a particular convergence test for their value iteration experiments. The problem is not periodic and includes a discount factor, and we therefore we used a standard convergence test for the value function \parencite{sutton_reinforcement_2018} as set out in Appendix \ref{appendix:a:desc}.

\textcite{de_moor_reward_2022} considered products with a maximum useful life $m$ of two, three, four or five periods, and evaluated eight different experimental settings for each value of $m$. For a product with $m=2$, they found the optimal policy using value iteration, and used this as a benchmark for their deep reinforcement learning policies. For larger values of $m$, they instead used a heuristic policy as the benchmark on  grounds of computational feasibility. The experiments for each value of $m$ evaluate different combinations of lead time $L$, wastage cost $C_w$, and issuing policy. We demonstrate that, using JAX and a consumer-grade GPU, it is feasible to obtain the optimal policy for all of the experimental settings, up to and including a maximum useful life $m$ of five periods, and report the wall time required to run value iteration for each experiment. 

We compare the policy from value iteration with a standard base-stock policy, parameterised by order-up-to level $\texttt{S}$, such that the order quantity on day $t$, given total current stock (on hand and in transit) $I_t$ is:

\begin{equation}
A_t = \left[\texttt{S} - I_t\right]^+
\end{equation}

We evaluated the mean return for each value of $\texttt{S} \in \{0, ..., A_{\max}\}$ using the Optuna grid sampler. We compare the base-stock policy that achieves the highest mean return, characterised by parameter \texttt{S}$_{\text{best}}$, to the value iteration policy.

See Appendix \ref{appendix:a:desc} for additional information about Scenario A. 

\subsection{Results}

In Table \ref{tab:demoor:results} we present the wall time (WT) in seconds required to run value iteration and simulation optimization for each experimental setting. We also present the mean and standard deviation of the return obtained when using value iteration and best heuristic policies on 10,000 simulated rollouts, each 365 days long following a warm-up period of 100 days. 

The wall times reported in Table \ref{tab:demoor:results} show that, using our approach, the largest cases, with $m=5$ and $L=2$, can be solved using value iteration in under 20 minutes. The running time for simulation optimization is approximately constant at two seconds over the different problem sizes. This is consistent with the fact that the parameter space for the heuristic base-stock policy is the same for each experimental setting, because the maximum order quantity $A_{\max}$ does not change. 

As we would expect, we observe higher mean returns under a FIFO issuing policy than under a LIFO issuing policy and as $m$ increases due to lower wastage. The optimality gap is consistently higher for experiments with a longer lead time, suggesting that the age profile of the stock is more important when lead times are longer. 

See Appendix \ref{appendix:a:res} for the best parameters for the heuristic policy and KPIs for each experiment. 

\begin{table}[h!]
\centering
\footnotesize
\begin{tabular}{crrrrrrrrrrrr}
\toprule
 &  & \multicolumn{6}{r}{} & \multicolumn{2}{c}{\textbf{Value}} & \multicolumn{3}{c}{\textbf{Simulation}} \\
  &  & \multicolumn{6}{r}{} & \multicolumn{2}{c}{\textbf{iteration}} & \multicolumn{3}{c}{\textbf{optimization}} \\
 \cmidrule(lr){9-10} \cmidrule(lr){11-13}
$m$ & Exp & $L$ & $C_w$ & Issuing & $|\mathbb{S}|$ & $|\mathbb{A}|$ & $|\mathbb{\Omega}|$ & WT (s) & Return  & WT (s)  & Return  & Optimality \\
& & & & policy &  & &  &  & & & & gap (\%) \\
\midrule
2 & 1 & 1 & 7 & LIFO & 121 & 11 & 101 & 5 & -1,553 $\pm$ 61  & 2 & -1,565 $\pm$ 62  & 0.80 \\
 & 2 & 1 & 7 & FIFO & 121 & 11 & 101 & 4 & -1,457 $\pm$ 59  & 2 & -1,474 $\pm$ 56  & 1.20 \\
 & 3 & 1 & 10 & LIFO & 121 & 11 & 101 & 5 & -1,571 $\pm$ 61  & 2 & -1,581 $\pm$ 62  & 0.64 \\
 & 4 & 1 & 10 & FIFO & 121 & 11 & 101 & 5 & -1,463 $\pm$ 60  & 2 & -1,485 $\pm$ 61  & 1.46 \\
 & 5 & 2 & 7 & LIFO & 1,331 & 11 & 101 & 5 & -1,551 $\pm$ 62  & 2 & -1,590 $\pm$ 64  & 2.49 \\
 & 6 & 2 & 7 & FIFO & 1,331 & 11 & 101 & 5 & -1,461 $\pm$ 58  & 2 & -1,495 $\pm$ 60  & 2.31 \\
 & 7 & 2 & 10 & LIFO & 1,331 & 11 & 101 & 6 & -1,569 $\pm$ 61  & 2 & -1,606 $\pm$ 64  & 2.35 \\
 & 8 & 2 & 10 & FIFO & 1,331 & 11 & 101 & 5 & -1,469 $\pm$ 59  & 2 & -1,504 $\pm$ 60  & 2.41 \\
\midrule
3 & 1 & 1 & 7 & LIFO & 1,331 & 11 & 101 & 5 & -1,490 $\pm$ 58  & 2 & -1,500 $\pm$ 59  & 0.71 \\
 & 2 & 1 & 7 & FIFO & 1,331 & 11 & 101 & 5 & -1,424 $\pm$ 56  & 2 & -1,435 $\pm$ 52  & 0.74 \\
 & 3 & 1 & 10 & LIFO & 1,331 & 11 & 101 & 5 & -1,498 $\pm$ 61  & 2 & -1,512 $\pm$ 58  & 0.90 \\
 & 4 & 1 & 10 & FIFO & 1,331 & 11 & 101 & 5 & -1,425 $\pm$ 55  & 2 & -1,436 $\pm$ 52  & 0.82 \\
 & 5 & 2 & 7 & LIFO & 14,641 & 11 & 101 & 13 & -1,513 $\pm$ 61  & 2 & -1,533 $\pm$ 61  & 1.32 \\
 & 6 & 2 & 7 & FIFO & 14,641 & 11 & 101 & 13 & -1,435 $\pm$ 56  & 2 & -1,456 $\pm$ 58  & 1.42 \\
 & 7 & 2 & 10 & LIFO & 14,641 & 11 & 101 & 13 & -1,526 $\pm$ 60  & 2 & -1,544 $\pm$ 61  & 1.16 \\
 & 8 & 2 & 10 & FIFO & 14,641 & 11 & 101 & 13 & -1,437 $\pm$ 56  & 2 & -1,457 $\pm$ 58  & 1.42 \\
\midrule
4 & 1 & 1 & 7 & LIFO & 14,641 & 11 & 101 & 14 & -1,459 $\pm$ 56  & 2 & -1,476 $\pm$ 54  & 1.15 \\
 & 2 & 1 & 7 & FIFO & 14,641 & 11 & 101 & 14 & -1,422 $\pm$ 56  & 2 & -1,430 $\pm$ 52  & 0.54 \\
 & 3 & 1 & 10 & LIFO & 14,641 & 11 & 101 & 14 & -1,465 $\pm$ 56  & 2 & -1,481 $\pm$ 60  & 1.08 \\
 & 4 & 1 & 10 & FIFO & 14,641 & 11 & 101 & 14 & -1,422 $\pm$ 56  & 2 & -1,430 $\pm$ 52  & 0.54 \\
 & 5 & 2 & 7 & LIFO & 161,051 & 11 & 101 & 111 & -1,480 $\pm$ 59  & 2 & -1,496 $\pm$ 59  & 1.07 \\
 & 6 & 2 & 7 & FIFO & 161,051 & 11 & 101 & 110 & -1,432 $\pm$ 55  & 2 & -1,453 $\pm$ 58  & 1.44 \\
 & 7 & 2 & 10 & LIFO & 161,051 & 11 & 101 & 110 & -1,489 $\pm$ 59  & 2 & -1,505 $\pm$ 58  & 1.07 \\
 & 8 & 2 & 10 & FIFO & 161,051 & 11 & 101 & 109 & -1,432 $\pm$ 55  & 2 & -1,453 $\pm$ 58  & 1.44 \\
\midrule
5 & 1 & 1 & 7 & LIFO & 161,051 & 11 & 101 & 114 & -1,443 $\pm$ 55  & 2 & -1,454 $\pm$ 55  & 0.73 \\
 & 2 & 1 & 7 & FIFO & 161,051 & 11 & 101 & 113 & -1,422 $\pm$ 56  & 2 & -1,430 $\pm$ 52  & 0.54 \\
 & 3 & 1 & 10 & LIFO & 161,051 & 11 & 101 & 114 & -1,446 $\pm$ 56  & 2 & -1,460 $\pm$ 55  & 0.94 \\
 & 4 & 1 & 10 & FIFO & 161,051 & 11 & 101 & 114 & -1,422 $\pm$ 56  & 2 & -1,430 $\pm$ 52  & 0.54 \\
 & 5 & 2 & 7 & LIFO & 1,771,561 & 11 & 101 & 1,191 & -1,463 $\pm$ 58  & 2 & -1,480 $\pm$ 60  & 1.22 \\
 & 6 & 2 & 7 & FIFO & 1,771,561 & 11 & 101 & 1,185 & -1,432 $\pm$ 55  & 2 & -1,453 $\pm$ 58  & 1.44 \\
 & 7 & 2 & 10 & LIFO & 1,771,561 & 11 & 101 & 1,188 & -1,467 $\pm$ 58  & 2 & -1,484 $\pm$ 59  & 1.15 \\
 & 8 & 2 & 10 & FIFO & 1,771,561 & 11 & 101 & 1,190 & -1,432 $\pm$ 55  & 2 & -1,453 $\pm$ 58  & 1.44 \\
\bottomrule
\end{tabular}
\caption{\label{tab:demoor:results}Our results on Scenario A for all of the experimental settings from \textcite{de_moor_reward_2022}. The longest wall times, for value iteration when $m=5$ and $L=2$, are approximately 20 minutes. Value iteration was considered intractable for experiments where $m > 2$ in the original study.}
\end{table}

\section{Scenario B: substitution between two perishable products} \label{s:hendrix}

\subsection{Problem description}

\textcite{hendrix_computing_2019} applied value iteration and simulation optimization to fit replenishment policies for two perishable inventory problems: a single-product scenario that is similar to Scenario A and a scenario with two products and the potential for substitution which we consider here as Scenario B. In Scenario B we manage two perishable products, product A and product B, with the same fixed, known useful life $m$. Some customers who want product B are willing to accept product A instead if product B is out of stock. The lead time $L = 1$ and therefore there is no in transit component to the state. 

At the start of each day $t$, the agent observes state $S_t$, the current inventory of each product in stock split by remaining useful life, and places a replenishment order. The action consists of two elements, one order for each product: $A_t = [A^a_t, A^b_t]$ where $A^a_t \in \{0, 1, ..., A^a_{\max}\}$ and $A^b_t \in \{0, 1, ..., A^b_{\max}\}$. Demand for day $t$ is sampled from independent Poisson distributions for each product, parameterised respectively by mean demand $\mu^a$ and $\mu^b$, and is initially filled for each product independently using a FIFO issuing policy. Some customers with unmet demand for product B may be willing to accept product A instead. The substitution demand is sampled from a binomial distribution, with a probability of accepting substitution $\rho$ and a number of trials equal to the unmet demand for product B. After demand for product A has been filled as far as possible, demand for product B willing to accept product A is filled by any remaining units of product A using a FIFO issuing policy. At the end of the day, the state is updated to reflect the ageing of stock, and the reward, $R_{t+1}$ is calculated. The reward function comprises revenue per unit sold ($C_r^a, C_r^b$) and variable order cost ($C_v^a, C_v^b$) for each product. The order placed on day $t$ is received immediately prior to the start of day $t+1$ and is included in the stock element of state $S_{t+1}$. 

The daily demand and willingness to accept substitution are both stochastic. We capture the effect of both by considering the stochastic element in the transition to be the number of units issued for each product type: $H^a$ and $H^b$. The state transition and the reward are deterministic given a state-action pair and the number of units issued of product A and of product B. The set of possible realisations of the stochastic elements is:

\begin{align}
\mathbb{\Omega} =  \{(h^a, h^b)\} \quad \quad & h^a \in \{0, 1, ..., H^a_{\max}=mA^a_{\max}\} \\
& h^b \in \{0, 1, ..., H^b_{\max}=mA^b_{\max}\} \nonumber 
\end{align}

The initial value function, $V_0(s)$, is set to the expected sales revenue for state $s$ with $I^a$ units of product A and $I^b$ units of product B in stock. This is an infinite horizon problem with no discount factor, and therefore we used the convergence test specified in \textcite{hendrix_computing_2019}, which stops value iteration when the value of each state is changing by approximately the same amount on each iteration.  If the value of every state is changing by the same amount there will be no further changes to the best action for each state, indicating a stable estimate of the optimal policy. 

\textcite{hendrix_computing_2019} considered products with a maximum useful life $m$ of two and three periods and evaluated two experimental settings for each value of $m$. For experiments 1 and 2 with $m=3$, where the maximum order quantities were set based on the newsvendor model, they reported that it was not possible to complete value iteration within one week. They therefore repeated the two experiments for $m=3$ with lower values of $A^a_{\max}$ and $A^b_{\max}$. With this adjustment, one of the cases could be completed within 80 hours, while the other could still not be solved within a week. We demonstrate that using JAX and a consumer-grade GPU it is feasible to obtain the optimal policy for all of these settings and report the wall time required to run value iteration for each experiment. Additionally, to investigate how our method can benefit from more powerful GPUs, and how it scales to multiple GPUs, we report the wall times for running the largest problem on one, two and four Nvidia A100 40GB GPUs. 

Separately, we consider the experimental settings used by \textcite{ortega_cuda_2019} to evaluate their GPU-accelerated method, in which value iteration was always run for 100 iterations instead of to convergence. The different experiments evaluate mean daily demands between five and seven with maximum order quantities based on the newsvendor model.  

In each case, we compare the policy from value iteration with the modified base-stock policy used by \textcite{hendrix_computing_2019}, based on the work of \textcite{haijema_improved_2019}, which has an order-up-to level parameter for each product: $\texttt{S}^a$ and $\texttt{S}^b$. The order quantity for each product is determined considering only the on hand inventory of that product and includes an adjustment for expected waste. The order quantity on day $t$, given total stock on hand $I^a_t$ and $I^a_t$ and stock that expires at the end of the current period $X^a_{1,t}$ and $X^b_{1,t}$, is:

\begin{equation}
A_t = \left[A^a_t, A^b_t \right] = \biggl[\Bigl[\texttt{S}^a - I^a_t + \bigl[X^a_{1,t} - \mu^a\bigr]^+\Bigr]^+, \Bigl[\texttt{S}^b - I^b_t + \bigl[X^b_{1,t} - \mu^b\bigr]^+\Bigr]^+\biggr]
\end{equation}

 There are two parameters, and we used Optuna's NSGAII sampler to search the parameter space $\texttt{S}^a \in \{0, 1, ..., \texttt{S}^a_{\max}=2A^a_{\max}\}$ and $\texttt{S}^b \in \{0, 1, ..., \texttt{S}^b_{\max}=2 A^b_{\max}\}$. We considered values of the order-up-to level up to twice the maximum order quantity used for value iteration because \textcite{hendrix_computing_2019} reported best values of \texttt{S} that were higher than the values of $A_{\max}$ specified for value iteration for some of their experiments. We compare the modified base-stock policy that achieved the highest mean return, characterised by the pair of parameters $\left(\texttt{S}^a, \texttt{S}^b\right)_{\text{best}}$, to the value iteration policy. 

See Appendix \ref{appendix:b:desc} for additional information about Scenario B.

\subsection{Results}

We present results for the experimental settings for the two product scenario from \textcite{hendrix_computing_2019} in Table \ref{tab:hendrixtwoprod:results}. The wall times in Table \ref{tab:hendrixtwoprod:results} show that, using our method, value iteration can be used to find the optimal policy for all four settings of the two product scenario with $m=3$ in under 3.2 hours. \textcite{hendrix_computing_2019} reported that, for $m=3$,  value iteration did not converge within a week for experiments 1, 2 and 3 using a MATLAB implementation and experiment 4 converged in 80 hours. Our implementation of experiment 4, running on a consumer-grade GPU, converges in just over two minutes: more than 2000${\times}$ faster.

We present results for the four experimental settings, P1 to P4, from \textcite{ortega_cuda_2019} in Table \ref{tab:ortega:results}. The wall times for our approach are at least six times faster than those reported by \textcite{ortega_cuda_2019} for all four settings. We cannot conclude on the relative performance of our method and the GPU-accelerated method from \textcite{ortega_cuda_2019} without running both implementations on the same hardware and accounting for the difference between up-front and JIT compilation. However, the results suggest that our method is at least competitive with a custom CUDA implementation of value iteration for the two product case while requiring less specialist knowledge of GPU programming.

Simulation optimization scales well to larger problems, with wall times less than one minute for all of the experimental settings. The optimality gap is never greater than 1\%, and reduces as both mean demand and the maximum useful life increase. This suggests that there is a limited advantage to making ordering decisions based on the stock of both products, compared to making independent decisions for each product using a simple heuristic policy, under the reward function and substitution process proposed by \textcite{hendrix_computing_2019}. 

\begin{figure}[ht]
\centering
    \includegraphics{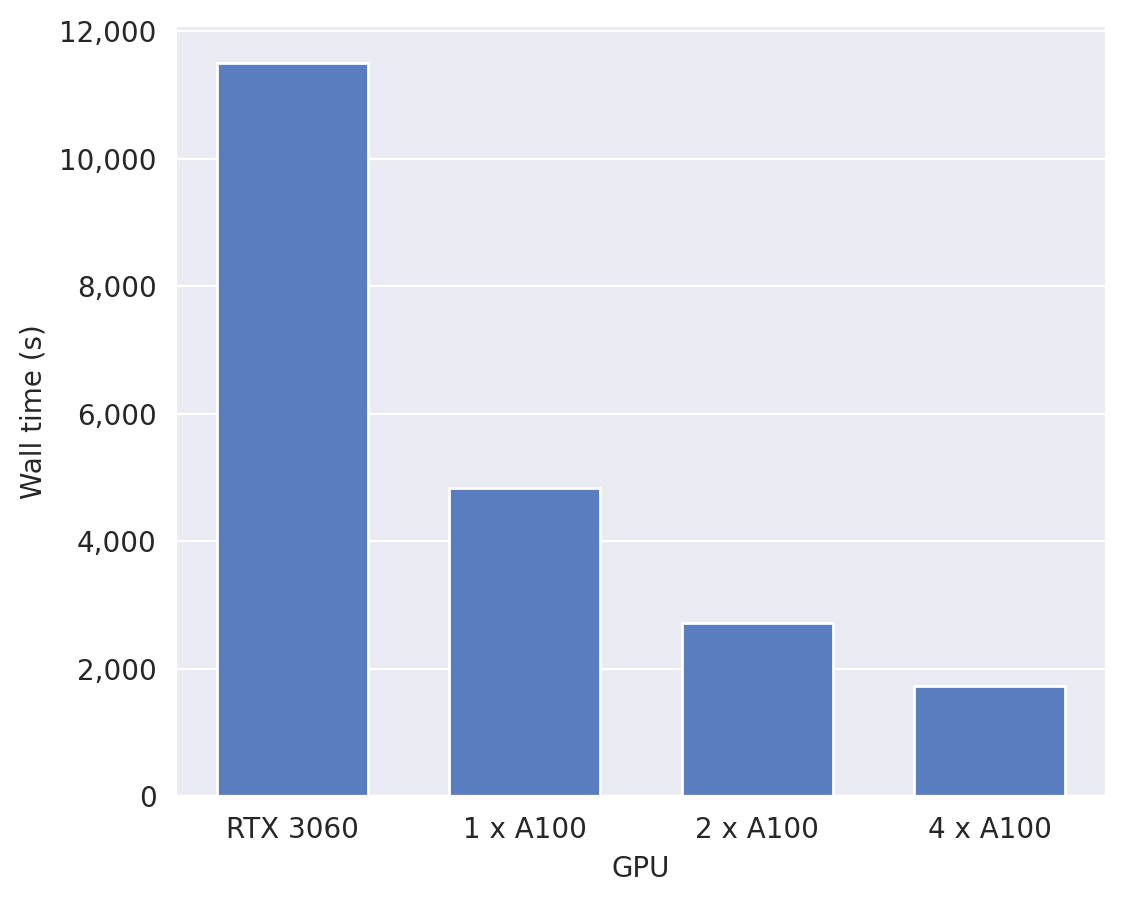}
    \caption{Wall times required to run value iteration for experiment 1 with $m=3$ for Scenario B using different GPUs. The Nvidia GeForce RTX 3060 is a consumer-grade GPU. The Nvidia A100 40GB is a data-centre grade GPU. JAX enables our method to be run on multiple identical GPUs without any code changes.}
    \label{fig:multigpu}
\end{figure}

Figure \ref{fig:multigpu} illustrates the clear benefits of both more powerful GPUs, and of using multiple GPUs. Using a single Nvidia A100 40GB GPU, experiment 1 when $m=3$ can be run in 4,838s: 2.4$\times$ faster than the Nvidia RTX 3060 in our local machine. The A100 40GB has more GPU RAM and more CUDA cores than the RTX 3060 \parencite{technicalcity_a100_nodate}, 40GB vs 16GB and 6,912 vs 3,584 respectively, which means that it can update the value function for a larger number of states simultaneously. Using two A100 40GB GPUs is 1.8$\times$ faster than one, and using four A100 40GB GPUs is 2.8$\times$ faster than one, demonstrating how the wall time can further reduced and how larger problems can be solved with additional computational resources using exactly the same code. 

See Appendix \ref{appendix:b:res} for the best parameters for the heuristic policy and KPIs for each experiment. 

\begin{table}[h!]
\centering
\footnotesize
\begin{tabular}{crrrrrrrrrrrrr}
\toprule
  & & \multicolumn{7}{r}{} & \multicolumn{2}{c}{\textbf{Value}} & \multicolumn{3}{c}{\textbf{Simulation}}  \\
    & & \multicolumn{7}{r}{} & \multicolumn{2}{c}{\textbf{iteration}} & \multicolumn{3}{c}{\textbf{optimization}} \\
 \cmidrule(lr){10-11} \cmidrule(lr){12-14}
 $m$ & Exp & $\mu^a$ & $\mu^b$ & $A^a_{\max}$ & $A^b_{\max}$ & $|\mathbb{S}|$ & $|\mathbb{A}|$ & $|\mathbb{\Omega}|$ & WT (s) & Return & WT (s) & Return & Optimality  \\
 &  &  &  &  &  &  &  &  &  &  &  &  & gap (\%) \\
\midrule
2 & 1 & 5 & 5 & 10 & 10 & 14,641 & 121 & 441 & 5 & 1,644 $\pm$ 33  & 24 & 1,632 $\pm$ 34  & 0.70 \\
 & 2 & 7 & 3 & 14 & 6 & 11,025 & 105 & 377 & 4 & 1,650 $\pm$ 33  & 23 & 1,639 $\pm$ 34  & 0.67 \\
\midrule
3 & 1 & 5 & 5 & 15 & 15 & 16,777,216 & 256 & 2,116 & 11,496 & 1,761 $\pm$ 32  & 33 & 1,758 $\pm$ 32  & 0.16 \\
 & 2 & 7 & 3 & 21 & 9 & 10,648,000 & 220 & 1,792 & 4,013 & 1,762 $\pm$ 32  & 44 & 1,759 $\pm$ 32  & 0.18 \\
 & 3 & 5 & 5 & 13 & 13 & 7,529,536 & 196 & 1,600 & 3,058 & 1,761 $\pm$ 32  & 32 & 1,758 $\pm$ 32  & 0.16 \\
 & 4 & 7 & 3 & 20 & 4 & 1,157,625 & 105 & 793 & 134 & 1,762 $\pm$ 32  & 43 & 1,759 $\pm$ 32  & 0.17 \\
\bottomrule
\end{tabular}
\caption{\label{tab:hendrixtwoprod:results}Our results on Scenario B for all of the experimental settings from \textcite{hendrix_computing_2019}. The longest wall time, for value iteration on experiment 1 when $m=3$, is approximately $3.2$ hours. Value iteration could not be completed within a week for experiments 1, 2 and 3, and required 80 hours for experiment 4, when $m=3$ in \textcite{hendrix_computing_2019}.}
\end{table}

\begin{table}[h!]
\centering
\footnotesize
\begin{tabular}{crrrrrrrrrrrrr}
\toprule
  & & \multicolumn{7}{r}{} & \multicolumn{2}{c}{\textbf{Value}} & \multicolumn{3}{c}{\textbf{Simulation}}  \\
    & & \multicolumn{7}{r}{} & \multicolumn{2}{c}{\textbf{iteration}} & \multicolumn{3}{c}{\textbf{optimization}}  \\
 \cmidrule(lr){10-11} \cmidrule(lr){12-14}
 $m$ & Exp & $\mu^a$ & $\mu^b$ & $A^a_{\max}$ & $A^b_{\max}$ & $|\mathbb{S}|$ & $|\mathbb{A}|$ & $|\mathbb{\Omega}|$ & WT (s) & Return & WT (s) & Return & Optimality\\
 &  &  &  &  &  &  & & &  &  &  &  &  gap (\%) \\
\midrule
2 & P1 & 5 & 5 & 10 & 10 & 14,641 & 121 & 441 & 11 & 1,644 $\pm$ 33  & 25 & 1,632 $\pm$ 34  & 0.70 \\
 & P2 & 5 & 6 & 10 & 12 & 20,449 & 143 & 525 & 18 & 1,826 $\pm$ 35  & 31 & 1,816 $\pm$ 34  & 0.58 \\
 & P3 & 6 & 6 & 12 & 12 & 28,561 & 169 & 625 & 27 & 2,011 $\pm$ 36  & 31 & 2,000 $\pm$ 37  & 0.55 \\
 & P4 & 7 & 7 & 13 & 13 & 38,416 & 196 & 729 & 56 & 2,379 $\pm$ 39  & 29 & 2,368 $\pm$ 40  & 0.46 \\
\bottomrule
\end{tabular}
\caption{\label{tab:ortega:results}Our results on Scenario B for all of the experimental settings used by \textcite{ortega_cuda_2019} to test their GPU-accelerated approach. Value iteration was run for 100 iterations for each experiment instead of to convergence. Aside from this, experiment P1 is the same as experiment 1 with $m=2$ in Table \ref{tab:hendrixtwoprod:results}. The longest wall time, for value iteration on experiment P4, is approximately one minute. Value iteration was tractable for all of these settings in the original study but wall times using our method are at least six times faster than those reported by \textcite{ortega_cuda_2019}. This improvement may be at least partially attributable to hardware differences.}
\end{table}

\section{Scenario C: periodic demand and uncertain useful life on\\ arrival}\label{s:mirjalili}

\subsection{Problem description}

\textcite{mirjalili_data-driven_2022} described a perishable inventory problem that models the management of platelets in a hospital blood bank. There are two problem features not included in Scenarios A or B. Firstly, the demand is periodic, with an independent demand distribution for each day of the week. Secondly, the remaining useful life of products on arrival is uncertain, and this uncertainty may be exogenous or endogenous. The lead time $L$ is assumed to be zero and therefore there is no in transit component to the state. 

At the start of each day $t$ the agent observes state $S_t$, which specifies the day of the week and the current inventory in stock split by remaining useful life, and places a replenishment order $A_t \in \{0, 1, ..., A_{\max}\}$. This order is assumed to arrive instantly. The remaining useful life of the units on arrival is sampled from a multinomial distribution, the parameters of which may depend on the order quantity $A_t$. Demand for day $t$, $D_t$ is sampled from a truncated negative binomial distribution and filled from available stock using an oldest-unit first-out (OUFO) policy. At the end of the day, the state is updated to reflect the ageing of stock and the reward, $R_{t+1}$ is calculated. The reward function comprises four components: a holding cost per unit in stock at the end of the period ($C_h$), a shortage cost per unit of unmet demand ($C_s$), a wastage cost per unit that perishes at the end of the period ($C_w$) and a fixed ordering cost ($C_f$). Unlike Scenario A which also includes a holding cost, the holding cost is charged on units that expire at the end of the period. 

To reduce the number of possible states, we consider a limited case of this problem in which there is a maximum capacity of $A_{\max}$ for stock of each age. If, when an order is received, the sum of units in stock and units received with $k$ days of remaining useful life is greater than $A_{\max}$ then we assume the excess units are not accepted at delivery. The stock level with $k$ days of remaining useful life is therefore at most $A_{\max}$ when demand is sampled. This constraint is chosen to be consistent with the calculation of the total number of states in \textcite{mirjalili_data-driven_2022}, but there are alternative ways to apply the constraint (e.g. by discarding excess units at the end of each day along with wastage) and these may have different optimal policies. 

The stochastic elements in the transition are the daily demand, $D$, and the age profile of the units received to fill the order placed at the start of the day: $\underline{\text{Y}} = \left[Y_{m}, Y_{m-1}, ..., Y_{1}\right]$. The state transition and the reward are deterministic given a state-action pair, the daily demand, and the age profile of the units received. The set of possible realisations of the stochastic elements is:

\begin{align}
\mathbb{\Omega} = \{(d, \underline{\text{y}})\} \quad & d \in \{0, 1, ..., D_{\max}\} \\
&  y_i \in \{0, 1, ..., A_{\max}\}, \quad \forall i \in \{1, 2, ..., m\} \nonumber \\ 
& \sum_{i=1}^{m} y_i \leq A_{\max}\nonumber
\end{align}

The initial value function $V_0(s)$ was initialised at zero for every state. \textcite{mirjalili_data-driven_2022} did not specify a particular convergence test for his value iteration experiments. The problem is periodic, with a discount factor, and therefore we use a convergence test based on those described in \textcite{su_generalization_1972} which stops value iteration when the undiscounted change in the value function over a period (in this case, seven days) is approximately the same for every state. As in Scenario B, when the change in value is the approximately the same for every state there will be no further changes to the best action for every state, and hence, the estimated optimal policy is stable. 

\textcite{mirjalili_data-driven_2022} considered products with a maximum useful life of three, five or eight periods and stated that, due to the large state space, value iteration was intractable for this problem when $m \geq 5$. We were able to run value iteration when $m=5$, but not when $m=8$. For each value of $m$, \textcite{mirjalili_data-driven_2022} investigated five different settings for the distribution of useful life on arrival (one where the uncertainty was exogenous, and four where the uncertainty was endogenous). For each of these five settings, he evaluated six combinations of $C_f$ and $C_w$. Our objective was to demonstrate the feasibility of our approach and therefore, given the large number of experiments and long wall times when $m=5$, we ran two experiments for each value of $m$: one where the uncertainty in useful life on arrival was exogenous and one where it was endogenous. For $m=5$, we selected the settings from \textcite{mirjalili_data-driven_2022} that are based on real, observed data from a network of hospitals in Ontario, Canada instead of the additional settings created for sensitivity analysis. We report the wall time required to run value iteration for each experiment. 

We compare the policy from value iteration with an $(\texttt{s}, \texttt{S})$ policy. \textcite{mirjalili_data-driven_2022} did not fit heuristic policies, but suggested (\texttt{s}, \texttt{S}) as an example of a suitable heuristic policy for future work: the addition of a fixed ordering cost to the reward function means that it may be beneficial to include the reorder point parameter $\texttt{s}$ to avoid uneconomically small orders. We fit one pair of \texttt{s} and \texttt{S} for each day of the week, a total of 14 parameters. The order quantity on day $t$, given that the day of the week is $\tau$ and the total current stock on hand is $I_t$ is:

\begin{equation}
A_t = \begin{cases}
\left[\texttt{S}^{\tau} - I_t\right]^+ &\text{if $I_t \leq \texttt{s}^{\tau}$} \\
0 &\text{if $I_t > \texttt{s}^{\tau}$}
\end{cases}
\end{equation}

\noindent where $(\texttt{s}^{\tau}, \texttt{S}^{\tau})$ is the pair of parameters for day of the week $\tau$. 

We used Optuna's NSGAII sampler to search for combinations of $\texttt{s}^{\tau} \in \{0, 1, ..., \texttt{s}_{\max} = A_{\max}\}$ and $\texttt{S}^{\tau} \in \{0, 1, ..., \texttt{S}_{\max} = A_{\max}\} \; \forall \tau \in \{0, 1, .., 6\}$. This heuristic policy has a hard constraint that $\texttt{s}^{\tau} < \texttt{S}^{\tau} \; \forall \tau \in \{0, 1, .., 6\}$. Optuna does not support using hard constraints to restrict the search space, so we enforced the constraint by only allowing a non-zero order to be placed if the constraint was met. We compare the heuristic policy that achieved the highest mean return, characterised by parameters $\left(\left(\texttt{s}^0, \texttt{S}^0\right), ..., \left(\texttt{s}^6, \texttt{S}^6\right)\right)_{\text{best}}$, to the value iteration policy. 

See Appendix \ref{appendix:c:desc} for additional information about Scenario C.

\subsection{Results}

In Table \ref{tab:mirjalili:results} we present the results for the experimental settings from \textcite{mirjalili_data-driven_2022} that we have selected, two for each value of $m$. Using our method, it is possible to find the optimal policy using value iteration for $m=3$ and $m=5$ while accounting for uncertainty in useful life on arrival. The experiments where $m=5$ represent a real world problem: \textcite{mirjalili_data-driven_2022} fit the parameters for the demand distribution and distribution of useful life on arrival to observed data from a network of hospitals in Ontario, Canada. This is an important application of our value iteration method, demonstrating that it can be used to find optimal policies for problems of a realistic size. The alternative experimental settings evaluated by \textcite{mirjalili_data-driven_2022} but not repeated here have the same numbers of states, actions and possible random outcomes and therefore we would expect the wall times to be of a similar order as corresponding experiments reported in Table \ref{tab:mirjalili:results}. 

We were unable to complete value iteration when $m=8$. This problem has over 12.6 billion possible states, even with the restriction that we placed on the maximum stock holding of each age, and over 65 million possible random outcomes. It is not feasible to store the state array in the memory of our local machine, let alone run value iteration. However, we were able to fit a heuristic policy using simulation optimization in less than 20 minutes. 

The simulation optimization experiments for this scenario take longer than those of the other scenarios, between five and 20 minutes. This is due to the large number of possible combinations of parameters, because our heuristic policy require seven pairs of parameters $\left(\texttt{s},\texttt{S}\right)$, one for each weekday. Optuna does not support restricting the search space based on the constraint that $\texttt{s}^{\tau} < \texttt{S}^{\tau} \; \forall \tau \in \{0, 1, .., 6\}$ and therefore the size of the search space for each experiment is $(A_{\max} + 1)^{14} = 3.2 \times 10^{18}$, compared to only 11 possible parameters for the base-stock policy used for Scenario A and fewer than 1,000 possible combinations of parameters for even the largest scenarios from Scenario B. The heuristic policies perform well, with a maximum optimality gap of 1.22\%.
See Appendix \ref{appendix:c:res} for the best parameters for the heuristic policy and KPIs for each experiment. 

\begin{table}[h!]
\centering
\footnotesize
\begin{tabular}{crlrrrrrrrr}
\toprule
 &  & \multicolumn{4}{r}{} & \multicolumn{2}{c}{\textbf{Value}} & \multicolumn{3}{c}{\textbf{Simulation}} \\
  &  & \multicolumn{4}{r}{} & \multicolumn{2}{c}{\textbf{iteration}} & \multicolumn{3}{c}{\textbf{optimization}}  \\
   \cmidrule(lr){7-8} \cmidrule(lr){9-11}
 $m$ & Exp & Uncertainty  & $|\mathbb{S}|$ & $|\mathbb{A}|$ & $|\mathbb{\Omega}|$ & WT (s) & Mean return & WT (s) & Mean return & Optimality \\
  &  & in useful life &  &  &  &  &  &  &  & gap (\%)  \\
\midrule
3 & 1 & Exogenous & 3,087 & 21 & 37,191 & 15 & -410 $\pm$ 62  & 507 & -411 $\pm$ 63  & 0.26 \\
 & 2 & Endogenous & 3,087 & 21 & 37,191 & 17 & -349 $\pm$ 53  & 305 & -352 $\pm$ 55  & 1.04 \\
\midrule
5 & 1 & Exogenous & 1,361,367 & 21 & 1,115,730 & 178,078 & -312 $\pm$ 46  & 514 & -313 $\pm$ 50  & 0.34 \\
 & 2 & Endogenous & 1,361,367 & 21 & 1,115,730 & 178,023 & -312 $\pm$ 47  & 393 & -315 $\pm$ 46  & 1.22 \\
\midrule
8 & 1 & Exogenous & 12,607,619,787 & 21 & 65,270,205 & --- & ---& 618 & -293 $\pm$ 42  & --- \\
 & 2 & Endogenous & 12,607,619,787 & 21 & 65,270,205 & --- & ---  & 972 & -297 $\pm$ 43  & --- \\
\bottomrule
\end{tabular}
\caption{\label{tab:mirjalili:results}Our results on Scenario C for a subset of the experimental settings from \textcite{mirjalili_data-driven_2022}: two examples for each value of $m$. The longest wall time, for value iteration on experiment 1 when $m=5$, is approximately 49.5 hours. Value iteration was considered intractable for the experiments where $m \geq 5$ in the original study. We were able to use value iteration when $m=5$, but not when $m=8$.}
\end{table}

\begin{figure}[hb]
\centering
    \includegraphics[width=0.9\textwidth]{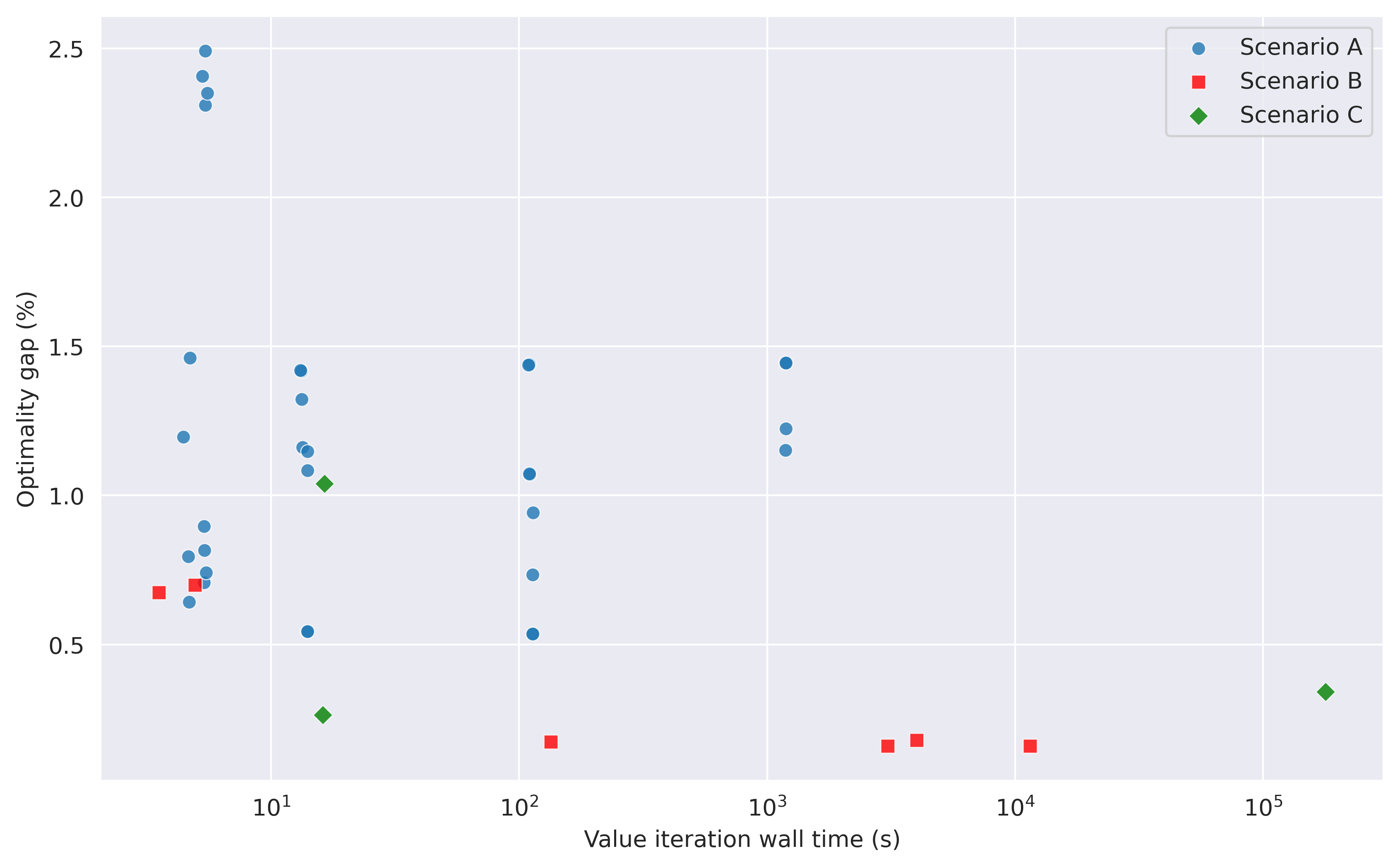}
    \caption{The optimality gap between the heuristic policy that achieved the highest mean return and the value iteration policy plotted against the wall time required to run value iteration for the experiments from Scenarios A, B and C.}
    \label{fig:optgap}
\end{figure}

In Figure \ref{fig:optgap} we draw together the results from Scenario C with those from the preceding scenarios, and plot the optimality gap between the heuristic policy that achieved the highest mean return and the value iteration policy against the wall time required for value iteration. 

\section{Discussion}

We have found JAX to provide an effective way to expand the scale of perishable inventory problems for which value iteration is tractable, using only consumer-grade hardware. Expanding the range of problems to which value iteration can be applied is not just a matter of considering settings with greater demand, more products, or products with a longer useful life. It also enables us to incorporate complexity that might otherwise be neglected due to its effect on the computational tractability such as substitution and endogenous uncertainty in the useful life of products on arrival. An ``optimal'' policy fit using value iteration is optimal for the situation as modelled, but may not perform well in practice if the model neglects challenging aspects of the real problem. For example, \textcite{mirjalili_data-driven_2022} reported large optimality gaps, with an average of 51\%, when policies obtained under the assumption that all stock arrived fresh were applied to a scenario with endogenous uncertainty in useful life. One avenue for future work is to consider scenarios that combine the more challenging elements: a multi-period lead time, substitution between multiple products, and uncertainty in the useful life on arrival which may all be relevant to managing blood product inventory in reality.

Our simulation optimization approach scales well to larger problems and large policy parameter spaces, and performs well relative to the optimal policies. One benefit of increasing the size of problems for which value iteration is tractable is being able to better understand how the relative of performance of heuristic and approximate policies scales with properties that influence the problem size. This may help with the development of new heuristics, and determining the utility of reinforcement learning and other approximate methods. The optimality gap in our experiments was never larger than 2.5\%, and in the experiments from Scenario A and Scenario B the optimality gap decreased as the demand and/or maximum useful life of the product increased. This is encouraging because it suggests that in some circumstances where the problem size remains too large for value iteration there may actually be little to gain by using the optimal policy over one of these heuristic policies. 

In our simulation optimization experiments we only used GPUs to run the simulated rollouts. The heuristic search methods for proposing the next sets of candidate parameters are CPU-based. We did not find Optuna's NSGAII sampler to be a bottleneck, but during preliminary experiments we found that some alternative methods took longer to propose the next set of candidate parameters than was required to evaluate them on simulated rollouts. In future work, the optimization process suggesting parameters could also be run on GPU similar to the work of \textcite{lau_hybrid_2016} and recent work using evolutionary strategies on GPUs to search for neural network parameters \parencite{lange_evosax_2022}. The gymnax-based simulators would also be well suited to ranking and selection methods because it would be straightforward to run a small number of rollouts for a large number of possible parameters in parallel and then, at a second stage, run a large number of rollouts for the most competitive parameters in parallel to obtain more accurate estimates of their performance. 

One of the main contributions of this work is to demonstrate an accessible way of using GPUs to accelerate value iteration and simulation optimization and therefore solve larger problems that are closer to those faced in reality. On the software side, we implemented our approach using the relatively high-level JAX API and relied on the XLA compiler to efficiently utilise GPU hardware. On the hardware side, we primarily report results on a consumer-grade GPU, and make available a Google Colab notebook so that our experiments can be reproduced at no cost using cloud-based computational resources. However, a significant strength of JAX is support for easily distributing a workload over multiple identical GPU devices using the pmap function transformation and we discuss in Section \ref{s:hendrix} how additional devices can be used to further reduce the wall time and potentially make even larger problems tractable. Modern cloud computing platforms provide on-demand access to data-centre grade GPUs, including the A100 40GB GPU we used to run the scaling experiments in Section \ref{s:hendrix}. At the time of writing in February 2023, a single A100 40GB GPU is available on-demand for \$3.67 per hour and four A100 40GB GPUs are available for \$14.68 per hour through Google Cloud Platform \parencite{karayev_cloud_2023}. This may provide a cost-effective way for research teams without access to local high-performance computing resources to investigate problems that are too large for freely available or consumer-grade GPU hardware. 

For some cases it may be possible to further reduce the wall time for value iteration on the same hardware by using single-precision numbers instead of double-precision numbers. We experienced instabilities in convergence during preliminary experiments with a large number of iterations which we resolved by changing from JAX's default single-precision to double-precision. The additional precision comes with a performance cost. Experiment 1 when $m=5$ for Scenario C has the longest wall time at double-precision: 49.5 hours. This can be reduced to 23.5 hours when using single-precision, over twice as fast at single-precision, and the final policy is the same using both approaches. A similar reduction in wall time can be obtained for experiment 2 with $m=5$ for Scenario C, but in this case the final policies are not identical. The best order quantities for just six out of 1,115,730 states differ from the policy found at double-precision, and each by one unit. This suggests that there may be a significant potential benefit to using single-precision numbers for GPU-accelerated value iteration if the user is willing to tolerate potentially larger approximation errors. 

Future hardware development will make value iteration feasible for even larger problems. In addition to future generations of GPUs, one promising direction is field programmable gate arrays (FPGAs): integrated circuits that can be reprogrammed to customise the hardware to implement a specific algorithm, including value iteration \parencite{peri_hardware_2020}. Customising the hardware currently requires specialist knowledge but, just as machine learning frameworks and higher-level tools have made GPU programming more accessible, \textcite{peri_hardware_2020} suggests that FPGA compilers able to translate high level code into customised circuit designs may facilitate wider adoption. 

We have focused on perishable inventory management in this study, but our computational approach has much wider applicability. For each scenario we created a custom subclass of our base value iteration runner class and a custom subclass of the gymnax reinforcement learning environment as our simulator, each with methods to implement the scenario-specific logic. The same approach could be followed for any problem that can be modelled as an MDP. More broadly, we believe that JAX (and other software libraries originally developed to support deep learning including PyTorch and Tensorflow) offers an efficient way for researchers to   run large workloads in parallel on relatively affordable GPU hardware which may support research on a range of operational research problems.

\section{Conclusion}

JAX and similar software libraries provide a way for researchers without extensive experience of GPU programming to take advantage of the parallel processing capabilities of modern GPUs. In this study we have shown how a JAX-based approach can expand the range of perishable inventory management problems for which value iteration is tractable, using only consumer-grade hardware. We also created GPU-accelerated simulators for each scenario, in the form of JAX-based reinforcement learning environments, and demonstrated how these can be used to quickly fit the parameters of heuristic policies by simultaneously evaluating many sets of policy parameters on thousands of simulated rollouts in parallel. By reducing the wall time required to run value iteration and simulation optimization, these methods can support research into larger problems, both in terms of scale and the incorporation of aspects of reality that increase the computational complexity. The ability to find optimal policies using value iteration may provide a valuable benchmark for the evaluation of new heuristic and approximate methods, helping efforts to make the best use of scarce resources and reduce wastage of perishable inventory. This work is focused on perishable inventory management but we believe that our methods, and the underlying principle of using software developed by the machine learning community to parallelize workloads on GPU, may be applicable to many problems in operational research and have made our code publicly available to support future work. 

\section*{Author roles}

\textbf{Joseph Farrington:} Conceptualization, Methodology, Software, Investigation, Validation, Writing - Original Draft \textbf{Kezhi Li:} Supervision, Writing - Review \& Editing \textbf{Wai Keong Wong:} Supervision \textbf{Martin Utley:} Investigation, Supervision, Writing - Review \& Editing

\section*{Acknowledgements}

The authors are grateful to Professor Eligius Hendrix and Dr Mahdi Mirjalili for providing additional material that enabled us to test our implementation of the scenarios from their work. Any errors or differences are our responsibility alone. The authors would also like to thank Dr Thomas Monks for sharing his expertise on simulation optimization during the preliminary stages of this work. 

\section*{Funding Statement}
JF is supported by UKRI training grant EP/S021612/1, the CDT in AI-enabled Healthcare Systems. This study was supported by the Clinical and Research Informatics Unit at the National Institute for Health and Care Research University College London Hospitals Biomedical Research Centre. 

The authors acknowledge the use of the UCL Myriad High Performance Computing Facility (Myriad@UCL), and associated support services, in the completion of this work.

The sponsors of the research did not have a role in the study design, in the collection, analysis and interpretation of the data, in the writing of this report, or in the decision to submit this article for publication. 

For the purpose of open access, the authors have applied a Creative Commons Attribution (CC BY) licence to any Author Accepted Manuscript version arising.

\section*{Competing interests}
The authors have no competing interests to declare.

\printbibliography

\newpage

\appendix

\section{Additional information for Scenario A} \label{appendix:demoor}

\subsection{Scenario description}\label{appendix:a:desc}

In this appendix we recast the problem formulated by \textcite{de_moor_reward_2022} into a consistent notation used for all three of the scenarios. 

The state of the system, $S_t$, comprises two components: the orders in transit $\underline{\text{O}_t}$ and the units in stock $\underline{\text{X}_t}$:
\begin{align}
\underline{\text{O}_t} &= \left[O_{L-1,t} = A_{t-1}, O_{L-2,t}, ..., O_{1,t} \right] \\
\underline{\text{X}_t} &= \left[X_{m,t} = O_{1,t-1}, X_{m-1,t}, ..., X_{1,t}\right]
\end{align}

\noindent for a total of $(m + L - 1)$ elements, with lead time $L \geq 1$. The total number of possible states is therefore $(A_{\max} + 1)^{m + L -1}$. In the state $S_t = \left[\underline{\text{O}_t}, \underline{\text{X}_t}\right]$, the entries are ordered by ascending age: the first element is the order placed on day $t-1$ and the last element is the stock that will expire at the end of the current day. The total number of units in stock at the start of day $t$ is $X_t = \sum_{i=1}^{m} X_{i,t}$, the total number of units in transit at the start of day $t$ is $O_t = \sum_{i=1}^{L-1} O_{i,t}$. The total number of units in stock or in transit at the start of day $t$ is $I_t = X_t + O_t$. If $L=1$ there is no in transit component to the state, and the first element of $\underline{\text{X}_t}$ is $A_{t-1}$. In Table \ref{tab:de_moor:common_settings} we present the parameter values that are the same for all of the experiments for Scenario A. 

\begin{table}[h!]
\centering
\begin{tabular}{l C{1cm}C{1cm}C{1cm}C{1cm}C{1cm}C{1cm}C{1cm}C{1cm}C{1.5cm}}
\toprule
 & $D_{\max}$ & $A_{\max}$ & $C_v$  & $C_s$ & $C_h$ & $\mu$ & $\frac{\mu}{\sigma}$ & $\gamma$ & $\epsilon$ \\
\midrule
Value & 100 & 10 & 3 & 5 & 1 & 4 & 0.5 & 0.99 & $1\times10^{-4}$ \\
\bottomrule
\end{tabular}
\caption{\label{tab:de_moor:common_settings}Parameter values that are consistent for all of the experiments for Scenario A.}
\end{table}

Daily demand, the stochastic element in the transition, is modelled using a truncated gamma distribution. It does not depend on the state or the action. The demand for the product is discrete and the gamma distribution is continuous, so the probability that the daily demand is equal to $d \in \{0, 1, ..., D_{\max}\}$ is:

\begin{align}
\text{Prob}(\Omega=\omega|S=s,A=a) &= P(\Omega = d) \\
&= P(D=d) \nonumber \\
&= \begin{cases}
F(d+\frac{1}{2}; \mu, \frac{\mu}{\sigma}) - F(d-\frac{1}{2}; \mu, \frac{\mu}{\sigma}), & \text{if $d \in \{0, 1, ..., D_{\max}-1$\}} \nonumber \\
1 - F(D_{\max}-\frac{1}{2};\mu,\frac{\mu}{\sigma}), &  \text{if $d = D_{\max}$}
\end{cases}
\end{align}

\noindent where $F(x; \mu, \frac{\mu}{\sigma})$ is the cumulative distribution function of the gamma distribution parameterised by mean $\mu$ and coefficient of variation $\frac{\mu}{\sigma}$, and $F(x; \mu, \frac{\mu}{\sigma}) = 0$ when $x \leq 0$.

The reward function comprises four components: a holding cost per unit in stock at the end of the period ($C_h$), a variable ordering cost per unit ($C_v$), a shortage cost per unit of unmet demand ($C_s$) and a wastage cost per unit that perishes at the end of the period ($C_w$). The single-step reward after taking action $A_t$ in state $S_t$ with $\Omega_t = \left(D_t\right)$ is:

\begin{equation} \label{eq:demoor:reward}
R_{t+1} = -C_vA_t - C_h \left[X_t - D_t - W_t \right]^+ - C_s \left[D_t - X_t \right]^+ - C_w W_t
\end{equation}

\noindent where $W_t$ is the number of units that expire at the end of period $t$. 

Equations \ref{eq:demoor:fifo} and \ref{eq:demoor:lifo} set out how the number of expired units, $W_t$, is calculated and how the elements of $\underline{\text{X}_t}$ are updated when following a FIFO issuing policy and a LIFO issuing policy, respectively.

\begin{align} \label{eq:demoor:fifo}
W_t &= [X_{1,t} - D_t]^+ \\ \nonumber
X_{j,t+1} &= \left[X_{j+1,t} - \left[D_t - \sum_{k=1}^j X_{k,t} \right]^+\right]^+ \quad \forall j \in \{1, 2, ..., m-1\} \\
X_{m, t+1} &= O_{1,t} = A_{t-L+1} \nonumber
\end{align}

\begin{align} \label{eq:demoor:lifo}
W_t &= \left[X_{1,t} - \left[D_t - \sum^{m}_{k=2} X_{k,t} \right]^+\right]^+ \\ \nonumber
X_{j,t+1} &= \left[X_{j+1,t} - \left[D_t - \sum_{k=j+2}^{m} X_{k,t} \right]^+\right]^+ \quad \forall j \in \{1, 2, ..., m-1\} \\
X_{m, t+1} &= O_{1,t} = A_{t-L+1} \nonumber
\end{align}

The scenario is an infinite horizon MDP with a discount factor and no periodicity in the state space. We therefore used a standard convergence test for the value function \parencite{sutton_reinforcement_2018}, evaluating:
\begin{equation}
\max_{s \in \mathbb{S}} |V_i(s) - V_{i-1}(s)| < \epsilon
\end{equation}
after each iteration. The inequality tests for the convergence of the values themselves, and requires more iterations than the convergence tests used for the other scenarios which are testing for convergence of the change in value for each state. The test compares the current estimate of the value function with the estimate from the immediately preceding iteration and does not require previous checkpoints for evaluation. Therefore, to save storage space and writing time, we saved a checkpoint every 100 iterations.

\subsection{Additional results}\label{appendix:a:res}

We present additional results for Scenario A in Table \ref{tab:de_moor:kpis}: the order-up-to level parameter $\texttt{S}_{\text{best}}$ fit using simulation optimization and the mean and standard deviation of three KPIs calculated over 10,000 evaluation rollouts for each policy.

\begin{table}[h!]
\centering
\footnotesize
\begin{tabular}{crrrrrrrr}
\toprule
 \multicolumn{3}{c}{}&  \multicolumn{2}{c}{\textbf{Service level (\%)}} & \multicolumn{2}{c}{\textbf{Wastage (\%)}} & \multicolumn{2}{c}{\textbf{Holding (units)}} \\
    \cmidrule(lr){4-5} \cmidrule(lr){6-7} \cmidrule(lr){8-9}
 $m$ & Exp & $\texttt{S}_{\text{best}}$ & \multicolumn{1}{c}{VI} & \multicolumn{1}{c}{SO} & \multicolumn{1}{c}{VI}& \multicolumn{1}{c}{SO} & \multicolumn{1}{c}{VI} & \multicolumn{1}{c}{SO} \\
\midrule
2 & 1 & 5 & 61.0 $\pm$ 1.4 & 58.6 $\pm$ 1.3 & 2.4 $\pm$ 0.6 & 2.2 $\pm$ 0.6 & 0.2 $\pm$ 0.0 & 0.2 $\pm$ 0.0 \\
 & 2 & 7 & 72.7 $\pm$ 1.6 & 76.6 $\pm$ 1.5 & 0.7 $\pm$ 0.4 & 1.5 $\pm$ 0.5 & 0.5 $\pm$ 0.1 & 0.8 $\pm$ 0.1 \\
 & 3 & 5 & 61.0 $\pm$ 1.4 & 58.6 $\pm$ 1.3 & 2.4 $\pm$ 0.6 & 2.2 $\pm$ 0.6 & 0.2 $\pm$ 0.0 & 0.2 $\pm$ 0.0 \\
 & 4 & 6 & 71.7 $\pm$ 1.6 & 68.6 $\pm$ 1.5 & 0.7 $\pm$ 0.3 & 0.7 $\pm$ 0.3 & 0.5 $\pm$ 0.1 & 0.5 $\pm$ 0.0 \\
 & 5 & 7 & 61.0 $\pm$ 1.4 & 55.4 $\pm$ 1.3 & 2.4 $\pm$ 0.6 & 2.4 $\pm$ 0.7 & 0.2 $\pm$ 0.0 & 0.2 $\pm$ 0.0 \\
 & 6 & 9 & 73.5 $\pm$ 1.7 & 69.4 $\pm$ 1.5 & 0.9 $\pm$ 0.4 & 1.1 $\pm$ 0.4 & 0.6 $\pm$ 0.1 & 0.6 $\pm$ 0.1 \\
 & 7 & 7 & 61.0 $\pm$ 1.4 & 55.4 $\pm$ 1.3 & 2.4 $\pm$ 0.6 & 2.4 $\pm$ 0.7 & 0.2 $\pm$ 0.0 & 0.2 $\pm$ 0.0 \\
 & 8 & 9 & 72.3 $\pm$ 1.6 & 69.4 $\pm$ 1.5 & 0.8 $\pm$ 0.4 & 1.1 $\pm$ 0.4 & 0.6 $\pm$ 0.1 & 0.6 $\pm$ 0.1 \\
\midrule
3 & 1 & 6 & 69.5 $\pm$ 1.5 & 68.3 $\pm$ 1.4 & 1.3 $\pm$ 0.4 & 1.4 $\pm$ 0.4 & 0.5 $\pm$ 0.1 & 0.5 $\pm$ 0.0 \\
 & 2 & 8 & 79.3 $\pm$ 1.5 & 83.3 $\pm$ 1.4 & 0.1 $\pm$ 0.1 & 0.2 $\pm$ 0.2 & 0.9 $\pm$ 0.1 & 1.3 $\pm$ 0.1 \\
 & 3 & 6 & 65.2 $\pm$ 1.4 & 68.3 $\pm$ 1.4 & 0.7 $\pm$ 0.3 & 1.4 $\pm$ 0.4 & 0.4 $\pm$ 0.0 & 0.5 $\pm$ 0.0 \\
 & 4 & 8 & 79.3 $\pm$ 1.5 & 83.3 $\pm$ 1.4 & 0.1 $\pm$ 0.1 & 0.2 $\pm$ 0.2 & 0.9 $\pm$ 0.1 & 1.3 $\pm$ 0.1 \\
 & 5 & 8 & 65.6 $\pm$ 1.6 & 62.6 $\pm$ 1.4 & 1.7 $\pm$ 0.5 & 1.4 $\pm$ 0.5 & 0.3 $\pm$ 0.0 & 0.4 $\pm$ 0.0 \\
 & 6 & 10 & 78.1 $\pm$ 1.6 & 75.5 $\pm$ 1.5 & 0.1 $\pm$ 0.1 & 0.1 $\pm$ 0.1 & 0.9 $\pm$ 0.1 & 0.9 $\pm$ 0.1 \\
 & 7 & 8 & 65.6 $\pm$ 1.6 & 62.6 $\pm$ 1.4 & 1.7 $\pm$ 0.5 & 1.4 $\pm$ 0.5 & 0.3 $\pm$ 0.0 & 0.4 $\pm$ 0.0 \\
 & 8 & 10 & 77.9 $\pm$ 1.6 & 75.5 $\pm$ 1.5 & 0.1 $\pm$ 0.1 & 0.1 $\pm$ 0.1 & 0.9 $\pm$ 0.1 & 0.9 $\pm$ 0.1 \\
\midrule
4 & 1 & 7 & 74.4 $\pm$ 1.4 & 76.4 $\pm$ 1.5 & 0.7 $\pm$ 0.3 & 1.5 $\pm$ 0.4 & 0.7 $\pm$ 0.1 & 0.8 $\pm$ 0.1 \\
 & 2 & 8 & 79.3 $\pm$ 1.5 & 83.3 $\pm$ 1.4 & 0.0 $\pm$ 0.0 & 0.0 $\pm$ 0.0 & 0.9 $\pm$ 0.1 & 1.3 $\pm$ 0.1 \\
 & 3 & 6 & 73.7 $\pm$ 1.4 & 68.5 $\pm$ 1.4 & 0.6 $\pm$ 0.3 & 0.5 $\pm$ 0.3 & 0.7 $\pm$ 0.1 & 0.5 $\pm$ 0.1 \\
 & 4 & 8 & 79.3 $\pm$ 1.5 & 83.3 $\pm$ 1.4 & 0.0 $\pm$ 0.0 & 0.0 $\pm$ 0.0 & 0.9 $\pm$ 0.1 & 1.3 $\pm$ 0.1 \\
 & 5 & 9 & 69.5 $\pm$ 1.5 & 69.3 $\pm$ 1.5 & 1.0 $\pm$ 0.4 & 1.0 $\pm$ 0.4 & 0.5 $\pm$ 0.1 & 0.6 $\pm$ 0.1 \\
 & 6 & 10 & 78.9 $\pm$ 1.7 & 75.5 $\pm$ 1.5 & 0.0 $\pm$ 0.0 & 0.0 $\pm$ 0.0 & 1.0 $\pm$ 0.1 & 0.9 $\pm$ 0.1 \\
 & 7 & 9 & 68.7 $\pm$ 1.5 & 69.3 $\pm$ 1.5 & 0.9 $\pm$ 0.4 & 1.0 $\pm$ 0.4 & 0.5 $\pm$ 0.1 & 0.6 $\pm$ 0.1 \\
 & 8 & 10 & 78.9 $\pm$ 1.7 & 75.5 $\pm$ 1.5 & 0.0 $\pm$ 0.0 & 0.0 $\pm$ 0.0 & 1.0 $\pm$ 0.1 & 0.9 $\pm$ 0.1 \\
\midrule
5 & 1 & 7 & 76.3 $\pm$ 1.5 & 76.6 $\pm$ 1.5 & 0.4 $\pm$ 0.2 & 0.7 $\pm$ 0.3 & 0.8 $\pm$ 0.1 & 0.8 $\pm$ 0.1 \\
 & 2 & 8 & 79.3 $\pm$ 1.5 & 83.3 $\pm$ 1.4 & 0.0 $\pm$ 0.0 & 0.0 $\pm$ 0.0 & 0.9 $\pm$ 0.1 & 1.3 $\pm$ 0.1 \\
 & 3 & 7 & 75.6 $\pm$ 1.4 & 76.6 $\pm$ 1.5 & 0.3 $\pm$ 0.2 & 0.7 $\pm$ 0.3 & 0.8 $\pm$ 0.1 & 0.8 $\pm$ 0.1 \\
 & 4 & 8 & 79.3 $\pm$ 1.5 & 83.3 $\pm$ 1.4 & 0.0 $\pm$ 0.0 & 0.0 $\pm$ 0.0 & 0.9 $\pm$ 0.1 & 1.3 $\pm$ 0.1 \\
 & 5 & 9 & 71.9 $\pm$ 1.6 & 69.5 $\pm$ 1.5 & 0.6 $\pm$ 0.3 & 0.4 $\pm$ 0.3 & 0.6 $\pm$ 0.1 & 0.6 $\pm$ 0.1 \\
 & 6 & 10 & 78.9 $\pm$ 1.7 & 75.5 $\pm$ 1.5 & 0.0 $\pm$ 0.0 & 0.0 $\pm$ 0.0 & 1.0 $\pm$ 0.1 & 0.9 $\pm$ 0.1 \\
 & 7 & 9 & 71.5 $\pm$ 1.6 & 69.5 $\pm$ 1.5 & 0.5 $\pm$ 0.3 & 0.4 $\pm$ 0.3 & 0.6 $\pm$ 0.1 & 0.6 $\pm$ 0.1 \\
 & 8 & 10 & 78.9 $\pm$ 1.7 & 75.5 $\pm$ 1.5 & 0.0 $\pm$ 0.0 & 0.0 $\pm$ 0.0 & 1.0 $\pm$ 0.1 & 0.9 $\pm$ 0.1 \\
\bottomrule
\end{tabular}
\caption{\label{tab:de_moor:kpis}The best order-up-to level $\texttt{S}_{\text{best}}$, fit using simulation optimization, and KPIs for policies fit using value iteration (VI) and simulation optimization (SO) for all of the experimental settings for Scenario A from \textcite{de_moor_reward_2022}.}
\end{table}

\newpage

\section{Additional information for Scenario B} \label{appendix:hendrix2}

\subsection{Scenario description}\label{appendix:b:desc}

In this appendix we recast the problem formulated by \textcite{hendrix_computing_2019} into a consistent notation used for all three of the scenarios. 

The state of the environment, $S_t$ comprises two components, one for each product type. In the combined state $S_t = \left[\underline{\text{X}^a_t}, \underline{\text{X}^b_t}\right]$, the elements in each component are ordered by ascending age: 
\begin{align}
\underline{\text{X}^a_t} &= [X^a_{m,t} = A^a_{t-1}, X^a_{m-1,t}, ..., X^a_{1,t}]\\
\underline{\text{X}^b_t} &= [X^b_{m,t} = A^b_{t-1}, X^b_{m-1,t}, ..., X^b_{1,t}]
\end{align}

\noindent for a total number of $2m$ elements. The total number of possible states is therefore $\left(A_{\max}^a + 1 \right)^m + \left(A_{\max}^b + 1\right)^m$. The total number of units in stock at the start of period $t$ is $I^a_t = X^a_t = \sum_{i=1}^{m}X^a_{i,t}$ for product A and $I^b_t = X^b_t = \sum_{i=1}^{m}X^b_{i,t}$ for product B. In Table \ref{tab:hendrix_two_product:common_settings} we present the parameter values that are the same for all of the experiments for Scenario B. 

\begin{table}[h!]
\centering
\begin{tabular}{lC{1cm}C{1cm}C{1cm}C{1cm}C{1cm}C{1cm}C{1.5cm}}
\toprule
& $C^a_v$ & $C^b_v$ & $C^a_r$ & $C^b_r$ & $\rho$ & $\gamma$ & $\epsilon$ \\
\midrule
Value & 0.5 & 0.5 & 1.0 & 1.0 & 0.5 & 1.0 & $1\times10^{-4}$ \\
\bottomrule
\end{tabular}
\caption{\label{tab:hendrix_two_product:common_settings}Parameter values that are consistent for all of the experiments for Scenario B.}
\end{table}

The stochastic element of the transition is the number of products of each type issued, ($H^a, H^b$). The number of units of product B issued only depends on the demand for product B and the total stock of product B, but the number of units of product A that are issued depends on the demand for product A, the total stock of product A, and any excess demand for product B for which the customer is willing to accept product A. 

 Let the demand for product A be $D^a$, the demand for product B be $D^b$, the excess demand for product B where the customer is willing to accept product A be $D^u$ and the total demand for product A including any substitution be $D^z = D^a + D^u$. To calculate the probability of a combination $\omega = (h^a, h^b)$ given a particular state $s$, we consider five possible cases:

\begin{align}
\text{Prob}\left(\Omega=\omega|S=s,A=a\right) &= P\left(\Omega = \left(h^a, h^b\right)|S=s\right)  \\
& = P\left(H^a = h^a, H^b = h^b|S=s\right) \nonumber \\
&= \begin{cases}
0, & \text{if } h^a > I^a  \text{ or } h^b > I^b \nonumber \\
P(D^a = h^a) P(D^b=h^b), \phantom{onetwothreefourfivesixse} & \text{if } h^a < I^a  \text{ and } h^b < I^b \\
P(D^a \geq I^a) P(D^b=h^b),& \text{if } h^a = I^a \text{ and } h^b < I^b \\
P(D^z = h^a|S=s)P(D^b \geq I^b),& \text{if } h^a < I^a \text{ and } h^b = I^b \\
P(D^z \geq I^a)P(D^b \geq I^b),& \text{if } h^a = I^a \text{ and } h^b = I^b \\
\end{cases} \\
&= 
\begin{cases}
0,& \text{if } h^a > I^a  \text{ or } h^b > I^b \nonumber \\
 P(h^a; \mu^a)P(h^b;\mu^b),& \text{if } h^a < I^a  \text{ and } h^b < I^b \\
\left[1 - F\left(I^a-1;\mu^a\right)\right]P(h^b;\mu^b),& \text{if } h^a = I^a \text{ and } h^b < I^b \\
P(D^z = h^a|S=s)\left[1 - \left(F(I^b-1; \mu^b\right) \right],& \text{if } h^a < I^a \text{ and } h^b = I^b \\
\left[1 - \sum_{d=0}^{I^a-1}P(D^z = d|S=s)\right] \left[1 - \left(F(I^b-1; \mu^b \right)\right],& \text{if } h^a = I^a \text{ and } h^b = I^b \\
\end{cases}
\end{align}

For the fourth and fifth cases there may be substitution, and therefore we need to consider the distribution of the total demand for product A and the distribution of the demand for substitution:

\begin{align}
P(D^z = d^z|S=s) & =  P(D^z = d^z|I^b = y) \nonumber \\ 
& = \sum_{k=0}^{d^z} P(D^a = k) P(D^u = d^z-k| I^b = y, D^b \geq y) \\
& = \sum_{k=0}^{d^z} P(k;\mu^a) P(D^u = d^z-k| I^b = y, D^b \geq y) \nonumber
\end{align}

\begin{align}
  P(D^u = d^u|I^b = y, D^b \geq y)&=
  \begin{cases}
    \sum_{c=0}^{\infty} P(D^b = c+y) (1-\rho)^c, & \text{if $d^u=0$}\\
    \sum_{c=d^u}^{\infty} P(D^b = c+y) P(d^u;c, \rho), & \text{if $d^u>0$}\\
  \end{cases} \\
  &=
  \begin{cases}
    \sum_{c=0}^{\infty} P(c+y; \mu^b) (1-\rho)^c,&\phantom{ac} \text{if $d^u=0$} \nonumber\\
    \sum_{c=d^u}^{\infty} P(c+y; \mu^b) P(d^u;c, \rho), &\phantom{ac} \text{if $d^u>0$}
  \end{cases}
\end{align}

\noindent where $P(x;c,\rho)$ is a binomial probability mass function representing the probability that there are $x$ units of excess demand for product B willing to accept product A out of a total of $c$ units of excess demand for product B and the probability of being willing to accept the substitution is $\rho$. $P(x; \mu^a)$ and $P(x; \mu^b)$ are the probability mass functions of independent Poisson distributions for the daily demand of product A and B parameterised by mean daily demands $\mu^a$ and $\mu^b$ respectively, and $F(x;\mu^a)$ and $F(x;\mu^b)$ are the corresponding cumulative distribution functions. 

We calculated the values of $P(D^u = d^u|I^b = y, D^b \geq y)$ and $P(D^z = d^z|S=s)$, for $d^u \in \{0, 1, ...,  D_{\max}\}$ and $d^z \in \{0, 1, ..., D_{\max}\}$, where $D_{\max} = \left(\left(m \max(A^a_{\max}, A^b_{\max})\right) + 2\right)$, at the start of value iteration following the MATLAB implementation of \textcite{hendrix_computing_2019}.

The reward function comprises two components which can be different for each product: a variable ordering cost per unit ($C^a_v$, $C^b_v$) and revenue per unit sold ($C^a_r$, $C^b_r$). The single step reward after taking action $A_t$ in state $S_t$ with ${\Omega}_t = \left(H^a_t, H^b_t\right)$ is:

\begin{equation} \label{eq:hendrix:oneproduct:reward}
R_{t+1} = - \left(C^a_v A^a_t + C^a_v A^b_t\right) +  \left(C^a_r H^a_t + C^b_r H^b_t\right)
\end{equation}

Equation \ref{eq:hendrix:twoproduct:fifo} shows how the elements of $\underline{\text{X}^a_t}$ and $\underline{\text{X}^b_t}$  are updated following a FIFO issuing policy. 

\begin{align} \label{eq:hendrix:twoproduct:fifo}
X^a_{j,t+1} &= X^a_{j+1,t} - \left[H^a_t - \sum_{k=1}^j X^a_{k,t} \right]^+ \quad \forall j \in \{1, 2, ..., m-1\} \\
X^a_{m, t+1} &= A^a_t \nonumber \\
X^b_{j,t+1} &= X^b_{j+1,t} - \left[H^b_t - \sum_{k=1}^j X^b_{k,t} \right]^+ \quad \forall j \in \{1, 2, ..., m-1\} \nonumber \\
X^b_{m, t+1} &= A^b_t \nonumber
\end{align}

The maximum order quantities for value iteration, $A^a_{\max}$ and $A^b_{\max}$ are calculated independently for each product following the newsvendor model \parencite{snyder_fundamentals_2019}: 
\begin{equation}\label{eq:hendrix:oneproduct:Omax}
A^k_{\max} = \Bigg \lceil F^{-1}\left(\frac{C^k_r - C^k_v}{C^k_r}; m\mu^k \right) \Bigg \rceil^+, \quad \forall k \in \{a, b\}
\end{equation}
where $F(x; m\mu^k)$ is the cumulative distribution function of a Poisson distribution parameterised by $m\mu^k$ and $\mu^k$ is the mean daily demand for product $k$. 

The maximum order quantities reported for experiments P1 to P4 in Table 3 of \textcite{ortega_cuda_2019} are not consistent with the number of states reported in that table. We have assumed that they instead represent the number of actions (one higher than the maximum order quantity, due to the possibility of ordering zero units) as this is consistent with the number of states reported, with Table 1 of \textcite{ortega_cuda_2019} and, for experiment P1, with the corresponding experiment in \textcite{hendrix_computing_2019}.

The initial estimate for the value function is the expected one-step ahead sales revenue:
\begin{equation}
V_0(s) = \sum^{I^a_t}_{h^a=0} \sum^{I^b_t}_{h^b=0} P(H^a=h^a, H^b=h^b|S=s) (h^aC^a_r + h^bC^b_r)
\end{equation}

We used the same convergence test as \textcite{hendrix_computing_2019}, evaluating:
\begin{equation}\label{eq:hendrix:oneproduct:convergence}
\max_{s \in \mathbb{S}}\left[V_i(s) - V_{i-1}(s)\right] - \min_{s \in \mathbb{S}}\left[V_i(s) - V_{i-1}(s)\right] < \epsilon
\end{equation}
after each iteration. The inequality tests for the convergence of the change in value for each state. When the value of each state is changing by the same amount, the best action for each state will not change and therefore the estimate of the optimal policy is stable. We saved a checkpoint after every iteration. 

\subsection{Additional results}\label{appendix:b:res}

We present additional results for the experimental settings from \textcite{hendrix_computing_2019} in Table \ref{tab:hendrixtwoprod:kpis} and for the experimental settings from \textcite{ortega_cuda_2019} in Table \ref{tab:hendrixtwoprod_ortega:kpis}. In each table we present the best combination of order-up-to level parameters $\left(\texttt{S}^a, \texttt{S}^b\right)_{\text{best}}$ fit using simulation optimization and the mean and standard deviation of three KPIs calculated over 10,000 evaluation rollouts for each policy. Demand for product B was considered to be satisfied for the purposes of calculating the service level if filled by product A when substitution was acceptable. 

The only difference between experiment 1 from \textcite{hendrix_computing_2019} and experiment P1 from \textcite{ortega_cuda_2019} is that value iteration was run for 100 iterations for experiment P1: more than were required for convergence. The best parameters for the modified base-stock policy and the KPIs from evaluating the policies are the same for these experiments, as we would expect. 

\begin{table}[h!]
\centering
\footnotesize
\begin{tabular}{crrrrrrrrr}
\toprule
\multicolumn{4}{c}{} & \multicolumn{2}{c}{\textbf{Service level (\%)}} & \multicolumn{2}{c}{\textbf{Wastage (\%)}} & \multicolumn{2}{c}{\textbf{Holding (units)}} \\
\cmidrule(lr){5-6} \cmidrule(lr){7-8} \cmidrule(lr){9-10}
$m$ & Exp & Product & $\texttt{S}_{\text{best}}$ & \multicolumn{1}{c}{VI} & \multicolumn{1}{c}{SO} & \multicolumn{1}{c}{VI} & \multicolumn{1}{c}{SO} & \multicolumn{1}{c}{VI} & \multicolumn{1}{c}{SO}  \\
\midrule
2 & 1 & A & 13 & 95.5 $\pm$ 0.8 & 95.2 $\pm$ 0.8 & 6.0 $\pm$ 1.0 & 6.3 $\pm$ 1.0 & 2.7 $\pm$ 0.1 & 2.7 $\pm$ 0.1 \\
 &  & B & 12 & 94.9 $\pm$ 0.8 & 95.5 $\pm$ 0.7 & 4.2 $\pm$ 0.8 & 5.3 $\pm$ 0.9 & 2.1 $\pm$ 0.1 & 2.3 $\pm$ 0.1 \\
  &  &  &  &  &  &  &  &  \\
 & 2 &  A & 18 & 96.9 $\pm$ 0.6 & 96.6 $\pm$ 0.6 & 4.2 $\pm$ 0.8 & 4.4 $\pm$ 0.7 & 3.7 $\pm$ 0.2 & 3.6 $\pm$ 0.2 \\
 &  &  B & 7 & 91.5 $\pm$ 1.1 & 92.5 $\pm$ 1.0 & 6.5 $\pm$ 1.2 & 8.3 $\pm$ 1.3 & 1.2 $\pm$ 0.1 & 1.3 $\pm$ 0.1 \\
\midrule
3 & 1 &  A & 15 & 98.3 $\pm$ 0.5 & 98.3 $\pm$ 0.5 & 2.2 $\pm$ 0.6 & 2.4 $\pm$ 0.6 & 4.9 $\pm$ 0.2 & 4.8 $\pm$ 0.2 \\
 &  &  B & 14 & 98.4 $\pm$ 0.4 & 98.4 $\pm$ 0.4 & 1.5 $\pm$ 0.5 & 1.7 $\pm$ 0.5 & 4.1 $\pm$ 0.2 & 4.1 $\pm$ 0.2 \\
  &  &  &  &  &  &  &  &  \\
 & 2 &  A & 21 & 99.1 $\pm$ 0.3 & 99.2 $\pm$ 0.3 & 1.2 $\pm$ 0.4 & 1.3 $\pm$ 0.4 & 6.7 $\pm$ 0.3 & 6.7 $\pm$ 0.2 \\
 &  &  B & 8 & 96.6 $\pm$ 0.7 & 96.1 $\pm$ 0.7 & 3.3 $\pm$ 0.9 & 3.2 $\pm$ 0.8 & 2.4 $\pm$ 0.1 & 2.3 $\pm$ 0.1 \\
  &  &  &  &  &  &  &  &  \\
 & 3 &  A & 15 & 98.3 $\pm$ 0.5 & 98.3 $\pm$ 0.5 & 2.2 $\pm$ 0.6 & 2.4 $\pm$ 0.6 & 4.9 $\pm$ 0.2 & 4.8 $\pm$ 0.2 \\
 &  &  B & 14 & 98.4 $\pm$ 0.4 & 98.4 $\pm$ 0.4 & 1.5 $\pm$ 0.5 & 1.7 $\pm$ 0.5 & 4.1 $\pm$ 0.2 & 4.1 $\pm$ 0.2 \\
  &  &  &  &  &  &  &  &  \\
 & 4 &  A & 21 & 99.1 $\pm$ 0.3 & 99.2 $\pm$ 0.3 & 1.2 $\pm$ 0.4 & 1.3 $\pm$ 0.4 & 6.6 $\pm$ 0.3 & 6.7 $\pm$ 0.2 \\
 &  & B & 8 & 96.6 $\pm$ 0.7 & 96.1 $\pm$ 0.7 & 3.3 $\pm$ 0.9 & 3.2 $\pm$ 0.8 & 2.4 $\pm$ 0.1 & 2.3 $\pm$ 0.1 \\
\bottomrule
\end{tabular}
\caption{\label{tab:hendrixtwoprod:kpis}The best combination of order-up-to levels $\left(\texttt{S}^a, \texttt{S}^b\right)_{\text{best}}$, fit using simulation optimization, and KPIs for policies fit using value iteration (VI) and simulation optimization (SO) for all of the experimental settings for Scenario B from \textcite{hendrix_computing_2019}}
\end{table}

\begin{table}[h!]
\centering
\footnotesize
\begin{tabular}{crrrrrrrrr}
\toprule
\multicolumn{4}{c}{} & \multicolumn{2}{c}{\textbf{Service level (\%)}} & \multicolumn{2}{c}{\textbf{Wastage (\%)}} & \multicolumn{2}{c}{\textbf{Holding (units)}} \\
\cmidrule(lr){5-6} \cmidrule(lr){7-8} \cmidrule(lr){9-10}
$m$ & Exp & Product & $\texttt{S}_{\text{best}}$ & \multicolumn{1}{c}{VI} & \multicolumn{1}{c}{SO} & \multicolumn{1}{c}{VI} & \multicolumn{1}{c}{SO} & \multicolumn{1}{c}{VI} & \multicolumn{1}{c}{SO}  \\
\midrule
2 & P1 &  A & 13 & 95.5 $\pm$ 0.8 & 95.2 $\pm$ 0.8 & 6.0 $\pm$ 1.0 & 6.3 $\pm$ 1.0 & 2.7 $\pm$ 0.1 & 2.7 $\pm$ 0.1 \\
 &  & B & 12 & 94.9 $\pm$ 0.8 & 95.5 $\pm$ 0.7 & 4.2 $\pm$ 0.8 & 5.3 $\pm$ 0.9 & 2.1 $\pm$ 0.1 & 2.3 $\pm$ 0.1 \\
   &  &  &  &  &  &  &  &  \\
 & P2 & A & 13 & 95.5 $\pm$ 0.8 & 95.1 $\pm$ 0.8 & 6.0 $\pm$ 1.0 & 6.2 $\pm$ 0.9 & 2.7 $\pm$ 0.1 & 2.6 $\pm$ 0.1 \\
 &  &  B & 14 & 95.9 $\pm$ 0.6 & 95.5 $\pm$ 0.7 & 3.5 $\pm$ 0.7 & 3.7 $\pm$ 0.7 & 2.5 $\pm$ 0.1 & 2.5 $\pm$ 0.1 \\
   &  &  &  &  &  &  &  &  \\
 & P3 & A & 16 & 96.2 $\pm$ 0.7 & 96.8 $\pm$ 0.6 & 4.9 $\pm$ 0.9 & 6.0 $\pm$ 0.9 & 3.2 $\pm$ 0.2 & 3.4 $\pm$ 0.1 \\
 &  & B & 14 & 95.9 $\pm$ 0.6 & 95.8 $\pm$ 0.6 & 3.5 $\pm$ 0.7 & 3.7 $\pm$ 0.7 & 2.5 $\pm$ 0.1 & 2.5 $\pm$ 0.1 \\
   &  &  &  &  &  &  &  &  \\
 & P4 & A & 18 & 96.8 $\pm$ 0.6 & 96.7 $\pm$ 0.6 & 4.2 $\pm$ 0.8 & 4.5 $\pm$ 0.8 & 3.7 $\pm$ 0.2 & 3.7 $\pm$ 0.2 \\
 &  & B & 17 & 96.5 $\pm$ 0.6 & 97.1 $\pm$ 0.5 & 2.9 $\pm$ 0.6 & 3.8 $\pm$ 0.7 & 2.9 $\pm$ 0.2 & 3.2 $\pm$ 0.1 \\
\bottomrule
\end{tabular}
\caption{\label{tab:hendrixtwoprod_ortega:kpis}The best combination of order-up-to levels $\left(\texttt{S}^a, \texttt{S}^b\right)_{\text{best}}$, fit using simulation optimization, and KPIs for policies fit using value iteration (VI) and simulation optimization (SO) for all of the experimental settings for Scenario B from \textcite{ortega_cuda_2019}}
\end{table}

\newpage

\section{Additional information for Scenario C} \label{appendix:mirjalili}

\subsection{Scenario description}\label{appendix:c:desc}

In this appendix we recast the problem formulated by \textcite{mirjalili_data-driven_2022} into a consistent notation used for all three of the scenarios. 

The state of the environment, $S_t$, comprises two components: $\tau \in \{0, 1, ..., 6\}$, representing the day of the week, and the units in stock at the start of the day $\underline{\text{X}_t} = \left[X_{m-1,t}, X_{m-2,t}, ..., X_{1, t}\right]$. The lead time, $L$, is always zero which means that the units ordered on day $t$ are received before any demand arises on day $t$. There are therefore only $m-1$ elements in $\underline{\text{X}_t}$ and a total of $m$ elements, including $\tau$, in $S_t$.  

In the previous problems, the maximum value of each element of $\underline{\text{X}_t}$ was $A_{\max}$, because all units arrived with the same remaining useful life. All units received in the same period would be in the same element of $\underline{\text{X}_t}$. In this scenario, the remaining useful life on arrival is stochastic and therefore, depending on the policy, a series of orders could be received such that an element of $\underline{\text{X}_t}$ would exceed $A_{\max}$. We assume there is a maximum capacity of $A_{\max}$ for stock of each possible value of remaining useful life. Units received in excess of this limit are not accepted at the point of delivery. The total number of possible states is therefore $7 \times \left(A_{\max} + 1\right)^{m-1}$. The entries in $\underline{\text{X}_t}$ are ordered by ascending age: the first element represents stock with $m-1$ days before expiry, and the last element is the stock that will expire at the end of day $t$. In Table \ref{tab:mirjalili:common_settings} we present the parameter values that are the same for all of the experiments for Scenario C. 

\begin{table}[h!]
\centering
\begin{tabular}{l C{1cm}C{1cm}C{1cm}C{1cm}C{1cm}C{1cm}C{1cm}C{1.5cm}}
\toprule
& $D_{\max}$ & $A_{\max}$ & $C_f$ & $C_h$ & $C_s$ & $C_w$ & $\gamma$ & $\epsilon$ \\
\midrule
Value & 20 & 20 & 10 & 1 & 20 & 5 & 0.95 & $1\times10^{-4}$ \\
\bottomrule
\end{tabular}
\caption{\label{tab:mirjalili:common_settings}Parameter values that are consistent for all of the experiments for Scenario C.}
\end{table}

The stochastic elements in the transition are the daily demand $D$, and the age profile of the units received: $\underline{\text{Y}} = \left[Y_{m}, Y_{m-1}, ..., Y_{1}\right]$.

The probability of a given random outcome $\omega$ is the product of the probability of the demand given the state, and the probability of receiving units with a specific age profile given the action: 

\begin{align}
    \text{Prob}\left(\Omega=\omega|S=s,A=a\right) &= 
    P\left(\Omega = \left(d, \underline{\text{y}}\right)|S = s, A = a\right) \\
    &= P\left(D = d, \underline{\text{Y}} = \underline{\text{y}}|S = s, A = a\right) \nonumber \\
&= P\left(D = d|S=s\right) P\left(\underline{\text{Y}} = \underline{\text{y}}| A = a\right) \nonumber
\end{align}

Demand is modelled by truncated negative binomial distributions, one for each day of the week. The demand distribution therefore only depends on the weekday element of the state. The negative binomial distribution models the number of failures, $x$, in a series of repeated Bernoulli trials before achieving a specified number of successes. The probability that daily demand is equal to $d$ on weekday $\tau$ is:

\begin{align} \label{eq:mirjalili:demand}
P(D=d|S=s) = \begin{cases}
    P(d;n^{\tau}, \delta^{\tau}), & \text{if } d \in \{0, 1, ..., D_{\max}-1\} \\
    1 - F(D_{\max}-1;n^{\tau}, \delta^{\tau}), & \text{if } d = D_{\max} \\
\end{cases}
\end{align}

\noindent where $P(x;n^{\tau}, \delta^{\tau})$ is the probability mass function of a negative binomial distribution parameterised by a target number of successes $n^{\tau}$ and a mean  $\delta^{\tau}$ for weekday $\tau$ and $F(x;n^{\tau}, \delta^{\tau})$ is the corresponding cumulative distribution function. The probability of success in an individual Bernoulli trial is $p^{\tau} = \frac{n^{\tau}}{n^{\tau} + \delta^{\tau}}$. The parameters for each day of the week are set out in Table \ref{tab:mirjalili:demand_params}. 

\begin{table}[h!]
\centering
\begin{tabular}{c C{0.75cm} C{0.75cm} C{0.75cm} C{0.75cm} C{0.75cm} C{0.75cm} C{0.75cm} }
\toprule 
$\tau$ & 0 & 1 & 2 & 3 & 4 & 5 & 6\\
\midrule
$n^{\tau}$ & 3.5 & 11.0 & 7.2 & 11.1 & 5.9 & 5.5 & 2.2\\
$\delta^{\tau}$ & 5.7 & 6.9 & 6.5 & 6.2 & 5.8 & 3.3 & 3.4\\
\bottomrule
\end{tabular}
\caption{\label{tab:mirjalili:demand_params}Parameters of the demand distribution for each weekday from Monday ($\tau=0$) to Sunday ($\tau=6$)}
\end{table}

The remaining useful life of units on arrival is modelled by a multinomial distribution with a number of trials equal to the order quantity $a$ and a number of events equal to the maximum useful life $m$. The probability mass function for the distribution is:

\begin{align} \label{eq:mirjalili:ageonarrival}
P(\underline{\text{Y}} = \underline{\text{y}}|A=a) &= P(Y_m = y_m, ..., Y_1 = y_1|A=a) \\
&= \begin{cases}
    \frac{a!}{y_m!y_{m-1}! ... y_1!} p_m(a)^{y_m}p_{m-1}(a)^{y_{m-1}}...p_1(a)^{y_1}, & \text{if $a = \sum_{i=1}^m y_i$} \\
    0, & \text{if $a \neq \sum_{i=1}^m y_i$} \nonumber \\
    \end{cases}
\end{align}

The parameters of the multinomial distribution are modelled by an affine function of the order quantity $a$: 

\begin{equation}
\log\left( \frac{p_k(a)}{p_1(a)} \right) = c_0^k + c^k_1 a, \quad \forall k \in \{2, 3, .., m\}
\end{equation}

If the distribution of remaining useful life on arrival does not depend on order quantity, and therefore the uncertainty is exogenous, $c_1^k = 0 \quad \forall k \in \{2, 3, .., m\}$. The values of $c_0^k$ and $c_1^k$ for our experiments are set out in Table \ref{tab:mirjalili:useful_life_params}. These represent a subset of the experiments run by \textcite{mirjalili_data-driven_2022}. The parameters for the two experiments where $m=5$  were determined by \textcite{mirjalili_data-driven_2022} by fitting multinomial logistic regression models to observed data from a hospital system in Ontario, Canada. 

\begin{table}[h!]
\footnotesize
\centering
\begin{tabular}{c r  r r r r r r r r r r r r r r r r}
\toprule
$m$ & Exp & $c_0^2$ & $c_0^3$  & $c_0^4$ & $c_0^5$ & $c_0^6$ & $c_0^6$ & $c_0^8$  & $c_1^2$ & $c_1^3$  & $c_1^4$  & $c_1^5$ & $c_1^6$ & $c_1^7$ & $c_1^8$\\
\midrule
3 & 1 &  1.0 & 0.5 & & &  &  & & & \\
& 2&  1.0 & 0.5 & & & & & &0.40 & 0.80 & & &   \\
\midrule
5& 1 &  1.6 & 2.6 & 2.8 & 1.6 & & & & & \\
& 2 &  1.9 & 3.1 & 3.1 & 2.5 & & & & -0.03 & -0.06 & -0.03 & -0.09 &  \\
\midrule
8& 1&  0.8 & 1.4 & 1.9 & 2.3 & 1.7 & 1.2 & 0.8  \\
& 2 &  0.8 & 1.4 & 1.9 & 2.3 & 1.7 & 1.2 & 0.8 & -0.03 & -0.04 & -0.05 & -0.06 & -0.07 & -0.08 & -0.09   \\
\midrule
\end{tabular}
\caption{\label{tab:mirjalili:useful_life_params}Parameters for the affine function used to model the parameters of the multinomial distribution of remaining useful life on arrival for each experiment. These are a subset of the experiments described by \textcite{mirjalili_data-driven_2022}}.
\end{table}

The reward function comprises four components: a holding cost per unit in stock at the end of the period ($C_h$), a shortage cost per unit of unmet demand ($C_s$), a wastage cost per unit that perishes at the end of the period ($C_w$) and a fixed ordering cost which is incurred when $A_t > 0$ ($C_f$). The single-step reward function after taking action $A_t$ in state $S_t = \left(\tau_t, \underline{\text{X}_t} \right)$, and observing $\Omega_t = (D_t, \underline{\text{Y}_t})$ is 

\begin{multline}
R_{t+1} = -C_f \mathbbm{1}_{A_t>0} - C_h\left[Y_{m,t} + \sum^{m-1}_{i=1} \min\left(X_{i,t} + Y_{i,t}, A_{\max}\right) - D_t\right]^+ \\ - C_s \left[D_t - Y_{m,t} - \sum^{m-1}_{i=1} \min\left(X_{i,t} + Y_{i,t}, A_{\max}\right)\right]^+ -  C_w \left[\min \left(X_{1,t} + Y_{1,t}, A_{\max}\right) - D_t\right]^+
\end{multline}

Equation \ref{eq:mirjalili:state_transition} shows how the elements of the state are updated, following a OUFO policy:

\begin{align} \label{eq:mirjalili:state_transition}
\tau_{t+1} &= (\tau_t + 1) \mod 7 \\ \nonumber
X_{j, t+1} &= \left[ \min \left(X_{j+1,t} + Y_{j+1,t}, A_{\max} \right) - \left[D_t - \sum^j_{k=1} \min\left(X_{k,t} + Y_{k,t}, A_{\max} \right)\right]^+\right]^+ \quad \forall j \in \{1, ... , m-2\} \\
X_{m-1, t+1} &= \left[Y_{m,t} - \left[D_t - \sum^{m-1}_{k=1} \min\left(X_{k,t} + Y_{k,t}, A_{\max}\right)\right]^+\right]^+ \nonumber
\end{align}

The scenario is an infinite horizon MDP with a discount factor and periodicity because the demand depends on the day of the week. We take advantage of the periodicity of the problem, and use a convergence test based on those described by \textcite{su_generalization_1972}. Performing this convergence test requires retaining at least the last seven (as this is the periodicity) estimates of the value function, and we can only test for convergence after we have run at least seven iterations. We tested the following inequality at the end of each iteration once $i \geq 7$:

\begin{align}
\Delta_{\text{max},i} &= \max_{s \in \mathbb{S}}\left[\sum_{j=0}^6 \frac{1}{\gamma^{i-j-1}} \left(V_{i-j}(s) - V_{i-j-1}(s)\right)\right]\\
\Delta_{\text{min},i} &= \min_{s \in \mathbb{S}} \left[\sum_{j=0}^6 \frac{1}{\gamma^{i-j-1}} \left(V_{i-j}(s) - V_{i-j-1}(s)\right) \right] \\
\Delta_{\text{max},i} - \Delta_{\text{min},i} &\leq 2 \epsilon \min \left[|\Delta_{\text{max},i}|, |\Delta_{\text{min},i}|\right]
\end{align}

When the inequality is met, the additional undiscounted reward being added to each state in one whole cycle (one week) is approximately the same, subject to our confidence level. In turn this means that for each weekday, every state is being increased by the same amount and therefore the best action for each state will not change. We therefore terminated value iteration when the inequality was met. If there were no discounting, and so $\gamma = 1$, the term in the square brackets would be equal to $V_i(s) - V_{i-7}(s)$: the total change in value from one cycle (in this case, one week). This convergence test relies on checkpoints from previous iterations, and we therefore saved a checkpoint every iteration. 

\subsection{Additional results}\label{appendix:c:res}

 In Table \ref{tab:mirjalili:heuristic_params} we present the best combination of parameters for the heuristic policy fit using simulation optimization. In Table \ref{tab:mirjalili:kpis} we present the mean and standard deviation of three KPIs calculated over 10,000 evaluation rollouts for each policy. For consistency with the calculation of the reward function in \textcite{mirjalili_data-driven_2022} the holding KPI includes the units that will expire at the end of the current day. These units are excluded from the calculation of the stock holding at the end of the day in the other scenarios 

\begin{table}[h!]
\centering
\footnotesize
\begin{tabular}{crrrrrrrrr}
\toprule
 &  &  & \multicolumn{7}{c}{\textbf{Weekday} $\tau$} \\
 \cmidrule(lr){4-10}
$m$ & Exp  & Parameter & 0 & 1 & 2 & 3 & 4 & 5 & 6 \\
\midrule
3 & 1 & $\texttt{S}_{\text{best}}$ & 13 & 12 & 14 & 11 & 11 & 8 & 7 \\
 &  & $\texttt{s}_{\text{best}}$ & 6 & 7 & 7 & 6 & 6 & 3 & 3 \\
  & &  &  &  &  &  &  &  &  \\
 & 2 & $\texttt{S}_{\text{best}}$ & 14 & 14 & 15 & 13 & 12 & 9 & 9 \\
 &  & $\texttt{s}_{\text{best}}$ & 7 & 7 & 7 & 7 & 6 & 3 & 4 \\
\midrule
5 & 1 & $\texttt{S}_{\text{best}}$ & 16 & 17 & 16 & 13 & 13 & 10 & 14 \\
 &  & $\texttt{s}_{\text{best}}$ & 7 & 8 & 8 & 7 & 7 & 3 & 3 \\
  & &  &  &  &  &  &  &  &  \\
 & 2 & $\texttt{S}_{\text{best}}$ & 17 & 16 & 16 & 13 & 13 & 11 & 14 \\
 &  & $\texttt{s}_{\text{best}}$ & 7 & 7 & 9 & 8 & 8 & 3 & 4 \\
\midrule
8 & 1 & $\texttt{S}_{\text{best}}$ & 19 & 15 & 18 & 18 & 14 & 13 & 16 \\
 &  & $\texttt{s}_{\text{best}}$ & 8 & 8 & 8 & 7 & 8 & 3 & 4 \\
  & &  &  &  &  &  &  &  &  \\
 & 2 & $\texttt{S}_{\text{best}}$ & 18 & 18 & 16 & 15 & 14 & 11 & 15 \\
 &  & $\texttt{s}_{\text{best}}$ & 9 & 8 & 9 & 7 & 7 & 3 & 4 \\
 \hline
\end{tabular}
\caption{\label{tab:mirjalili:heuristic_params}The best combination of parameters for the heuristic policy $\left(\left(\texttt{s}^0, \texttt{S}^0\right), ..., \left(\texttt{s}^6, \texttt{S}^6\right)\right)_{\text{best}}$ fit using simulation optimization for each of our experiments for Scenario C, a subset of the experiments run by \textcite{mirjalili_data-driven_2022}.}
\end{table}

\begin{table}[h!]
\centering
\footnotesize
\begin{tabular}{crrrrrrr}
\toprule
 &  & \multicolumn{2}{c}{\textbf{Service level (\%)}} & \multicolumn{2}{c}{\textbf{Wastage (\%)}} & \multicolumn{2}{c}{\textbf{Holding (units)}} \\
\cmidrule(lr){3-4} \cmidrule(lr){5-6} \cmidrule(lr){7-8}  
 $m$ & Exp & VI & SO & VI & SO & VI & SO \\
\midrule
3 & 1 & 95.3 $\pm$ 0.9 & 95.3 $\pm$ 0.9 & 12.6 $\pm$ 1.3 & 12.6 $\pm$ 1.4 & 4.9 $\pm$ 0.1 & 4.9 $\pm$ 0.1 \\
 & 2 & 96.6 $\pm$ 0.8 & 96.2 $\pm$ 0.8 & 7.0 $\pm$ 1.1 & 7.2 $\pm$ 1.1 & 5.8 $\pm$ 0.1 & 5.7 $\pm$ 0.1 \\
\midrule
5 & 1 & 97.4 $\pm$ 0.7 & 97.0 $\pm$ 0.7 & 3.2 $\pm$ 0.8 & 3.0 $\pm$ 0.7 & 6.8 $\pm$ 0.1 & 6.7 $\pm$ 0.1 \\
 & 2 & 97.4 $\pm$ 0.7 & 97.5 $\pm$ 0.7 & 3.1 $\pm$ 0.7 & 3.4 $\pm$ 0.8 & 6.8 $\pm$ 0.1 & 7.0 $\pm$ 0.2 \\
\midrule
8 & 1 & --- & 97.9 $\pm$ 0.6 & --- & 0.7 $\pm$ 0.3 & --- & 8.0 $\pm$ 0.2 \\
 & 2 & --- & 97.7 $\pm$ 0.6 & --- & 1.0 $\pm$ 0.4 & --- & 7.5 $\pm$ 0.2 \\
\bottomrule
\end{tabular}
\caption{\label{tab:mirjalili:kpis}KPIs for policies fit using value iteration (VI) and simulation optimization (SO) for each of our experiments for Scenario C, a subset of the experiments run by \textcite{mirjalili_data-driven_2022}. Value iteration was not feasible when $m=8$.}
\end{table}

\newpage

\section{Notation}\label{appendix:notation}

In Table \ref{tab:notation} we summarise the notation we have used to recast the problems described in Scenarios A, B and C. In Scenario B we use a superscript $a$ for product A and $b$ for product B if a variable is product-specific. In Scenario C we use a superscript $\tau$ (or weekday index from 0 to 6 representing Monday to Sunday, respectively) if a variable is weekday-specific. We drop the subscript $t$ in some contexts where elements from different days do not feature. 

\begin{table}[h!]
\centering
\footnotesize
\begin{tabular}{lll}
\toprule
Markov decision process & $\mathbb{S}$ & Set of possible states\\
& $S_t$ & State observed at the start of day $t$ \\
& $s$ & A specific element of $\mathbb{S}$\\
& $\mathbb{A}$ & Set of possible actions\\
& $A_t$ & Action taken at the start of day $t$ after observing $S_t$\\
& $a$ & A specific element of $\mathbb{A}$\\
& $\mathbb{\Psi}$ & Set of possible rewards \\
& $R_t$ & Reward received when state $S_t$ is observed\\
& $r$ &  A specific element of $\mathbb{\Psi}$\\
& $\mathbb{\Omega}$ & Set of possible realisations of stochastic elements in a transition\\
& $\Omega_t$ & Realisation of the stochastic elements in the transition between $S_t$ and $S_{t+1}$ \\
& $\omega$ & A specific element of $\mathbb{\Omega}$\\
& $\gamma$ & Discount factor\\
& $G_t$ & Return, the discounted sum of rewards received after taking action $A_t$\\
&$\pi(s)$ & Policy, a function mapping a state to an action\\
&$\pi^*(s)$ & Optimal policy, policy with the maximum expected return from every state\\
&$V^{\pi}(s)$ &Value function, expected return starting in state $s$ and following policy $\pi$\\
& $Q^{\pi}(s,a)$ &State-action value function, expected return taking action $a$ in state $s$ and \\ && following policy $\pi$ thereafter\\
& $T(s,a,\omega)$ & Deterministic transition function\\
\midrule
Reward function components & $C_v$ & Variable ordering cost per unit\\
& $C_f$ & Fixed ordering cost \\
& $C_w$ & Wastage cost per unit \\
& $C_s$ & Shortage cost per unit\\
& $C_h$ & Holding cost per unit\\
& $C_r$ & Revenue per unit \\
\midrule 
Value iteration & $\epsilon$ & Tolerance for convergence test\\
\midrule
Heuristic policy parameters & \texttt{S} & Order-up-to level \\
& \texttt{s} & Reorder point \\
\midrule
Inventory control & $L$ & Lead time\\
& $m$ & Maximum useful life\\
& $D_t$ & Demand on day $t$ \\
& $D_{\max}$ & Maximum daily demand \\
& $A_{\max}$ & Maximum daily order quantity \\
& \underline{X$_t$} & Vector of stock on hand at the start of day $t$, ordered by ascending age\\
& $X_{i,t}$ & Element of \underline{X$_t$} with $i$ days of remaining useful life at the start of day $t$\\
& $X_t$ & Total stock on hand at the start of day $t$\\
& $I_t$ & Total stock on hand and in transit at the start of day $t$\\
\midrule
Scenario A & \underline{O$_t$} & Vector of stock in transit at the start of day $t$, ordered by ascending age\\
& $O_{i,t}$ & Element of \underline{O$_t$} that will arrive in $i$ periods at the start of day $t$\\
& $O_t$ & Total stock in transit at the start of day $t$\\
& $W_t$ & Number of units that expire at the end of day $t$\\
& $\mu$ & Mean of the gamma distribution for daily demand \\
& $\frac{\mu}{\sigma}$ & Coefficient of variation of the gamma distribution for daily demand\\
\midrule
Scenario B & $H_t$ & Number of units of a product issued to fill demand arising on day $t$ \\
& $D_t^u$ & Excess demand for product B willing to accept product A on day $t$ \\
& $D_t^z$ & Total demand for product A, including any substitution, on day $t$ \\
& $\mu$ & Mean of the Poisson distribution for daily demand \\
& $\rho$ & Probability a customer is willing to accept product A if product B is out of stock \\
\midrule
Scenario C & $\tau_t$ & Day of the week for day $t$ \\
& \underline{Y$_t$} & Vector of stock received to fill order $A_t$, ordered by ascending age\\
& $Y_{i,t}$ & Element of \underline{Y$_t$} with $i$ periods of remaining useful life on arrival\\
& $Y_t$ & Total stock received on day $t$\\
& $n$ & Target number of successes for the negative binomial distribution for daily demand \\
& $\delta$ & Mean of the negative binomial distribution for daily demand \\
& $c_0^k$ & Log-odds ratio of receiving a unit with a remaining useful life of $k$ days versus\\ 
& & 1 day when the uncertainty is exogenous. \\
& $c_1^k$ & Increase in the log-odds ratio of receiving a unit with a remaining useful life of $k$ \\
& & days versus 1 day for each unit ordered when the uncertainty is endogenous\\ 
\bottomrule
\end{tabular}
\caption{\label{tab:notation}Summary of notation used in this study.}
\end{table}

\end{document}